\documentclass{article}

\PassOptionsToPackage{numbers, compress}{natbib}

\usepackage[preprint]{neurips_2024}
\usepackage[utf8]{inputenc} %
\usepackage[T1]{fontenc}    %
\usepackage{url}            %
\usepackage{booktabs}       %
\usepackage{amsfonts}       %
\usepackage{nicefrac}       %
\usepackage{microtype}      %
\usepackage{xcolor}         %

\title{Representing Animatable Avatar via \\Factorized Neural Fields}

\author{
  Chunjin Song$^{1}$, 
  Zhijie Wu$^{1}$, 
  Bastian Wandt$^{2}$, 
  Leonid Sigal$^{1}$, 
  Helge Rhodin$^{1,3}$\\
 $^1$ University of British Columbia, 
 $^2$ Linköping University,
 $^3$ Bielefeld University\\
  \texttt{\{chunjins, rhodin\}@cs.ubc.ca, bastian.wandt@liu.se} \\
}

\usepackage{multirow}
\usepackage{enumitem}

\usepackage{setspace}       %
\usepackage{hyperref}       %

\usepackage{times}
\usepackage{epsfig}

\usepackage{graphicx}

\usepackage{ragged2e}

\usepackage{tikz}
\usepackage{comment}
\usepackage{wrapfig} %
\usepackage{color}

\usepackage{mathtools}
\usepackage[accsupp]{axessibility}  %

\usepackage[utf8]{inputenc} %
\usepackage[T1]{fontenc}    %
\usepackage{xr-hyper}
\usepackage{url}            %
\usepackage{booktabs}       %
\usepackage{xcolor,colortbl}         %
\usepackage{caption}

\usepackage{multirow}
\usepackage{enumitem}

\usepackage[normalem]{ulem}

\usepackage{subcaption}
\usepackage{float}
\usepackage{lscape}     

\usepackage{xspace}
\usepackage{pifont}
\usepackage{comment}

\definecolor{turquoise}{cmyk}{0.65,0,0.1,0.1}
\definecolor{purple}{rgb}{0.65,0,0.65}
\definecolor{dark_green}{rgb}{0, 0.5, 0}
\definecolor{orange}{rgb}{0.8, 0.6, 0.2}
\definecolor{red}{rgb}{0.8, 0.2, 0.2}
\definecolor{blue}{rgb}{0, 0, 1}
\definecolor{brown}{rgb}{0.5, 0.16, 0.16}
\definecolor{black}{rgb}{0, 0, 0}

\definecolor{gray}{gray}{0.85}

\renewcommand{\paragraph}[1]{\vspace{0.3em}\noindent\textbf{#1}}

\newcommand{\bc}{\mathbf{c}}
\newcommand{\Color}{C}
\newcommand{\ray}{\mathbf{r}}
\newcommand{\Rays}{\mathfrak{R}}

\newif\ifdraft
\draftfalse %
\ifdraft

\newcommand{\hr}[1]{{\color{red}#1}}

\newcommand{\cj}[1]{{\color{blue}#1}}

\else

\newcommand{\hr}[1]{{#1}}

\newcommand{\cj}[1]{{#1}}

\fi

\begin{document}

\maketitle

\begin{abstract}
For reconstructing high-fidelity human 3D models from monocular videos, it is crucial to maintain consistent large-scale body shapes along with finely matched subtle wrinkles.
This paper explores the observation that 
\cj{
the per-frame rendering results can be factorized into a pose-independent component and a corresponding pose-dependent equivalent to facilitate frame consistency.
Pose adaptive textures can be further improved by restricting frequency bands of these two components.
In detail, pose-independent outputs are expected to be low-frequency, while high-frequency information is linked to pose-dependent factors.
}
We achieve a coherent preservation of both coarse body contours across the entire input video and fine-grained texture features that are time variant with a dual-branch network with distinct frequency components.
The first branch takes coordinates in canonical space as input, while the second branch additionally considers features outputted by the first branch and pose information of each frame. 
Our network integrates the information predicted by both branches and utilizes volume rendering to generate photo-realistic 3D human images.
Through experiments, we demonstrate that our network surpasses \cj{the neural radiance fields (NeRF) based} state-of-the-art methods in preserving high-frequency details and ensuring consistent body contours.

\end{abstract}
\section{Introduction}
\label{Sec:intro}

Neural body models now yield personalized, almost photorealistic 3D human avatars even from a single 2D video~\cite{weng2022humannerf,jiang2022neuman, yu2023monohuman, guo2023vid2avatar}. 
A widespread method is learning a Neural Radiance Field (NeRF)~\cite{su2021nerf,su2022danbo,weng2022humannerf} model in the observation space by conditioning the underlying neural network on the input pose. 
In detail, existing models impose constraints by learning the neural field relative to the skeleton and restricting pose-dependent changes to be local through a GNN~\cite{noguchi2021neural,su2021nerf,su2023npc,song2024pose}.
However, they have the risk of overfitting since a high capacity model could simply learn to render the training views without reconstructing a proper 3D shape. 
In turn, they are typically parameterized to smooth over high frequency details leading to artifacts in shape and texture, even when explicitly choosing a frequency-based representation~\cite{song2024pose}.

\begin{figure*}[t!]
    \centering
    \includegraphics[width=0.95\linewidth,trim={0 6cm 0cm 0cm},clip]{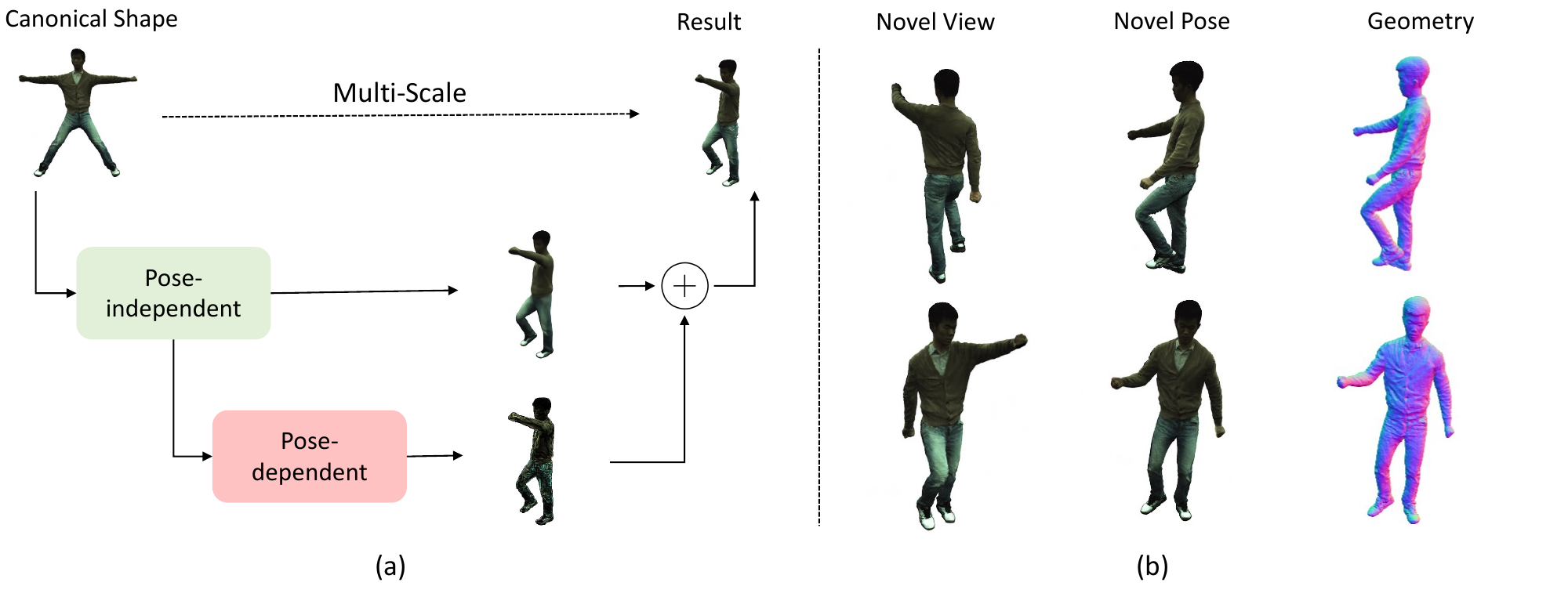}
    \vspace{-0.8em}
    \caption{
    \cj{
    \textbf{Motivation illustration.} (a) 
    In canonical space, we separate the per-frame rendering output into a pose-independent component and its pose-dependent equivalent.
    These two components are modeled with distinct frequency bands, thus yielding smooth base outputs and corresponding high-frequency residuals (see Fig.~\ref{fig:mocap-lf_hf} for details). The residual image here is amplified for better visualization.
    (b) Our frequency-aware factorized strategy improves the state-of-the-art methods in novel view synthesis, novel pose rendering and human shape reconstruction.
    }
    }
    \vspace{-1.2em}
    \label{fig:teaser}
\end{figure*}

The use of a canonical reference frame in which the shape and appearance of the character is defined statically, e.g. in a T-pose, is an alternative direction~\cite{jiang2022neuman,li2022tava,wang2022arah}. 
A 3D model can be learned from video by deforming the single static model to the pose visible in every frame with an explicit surface mesh~\cite{xu2018monoperfcap}, Gaussian representations~\cite{rhodin2015versatile,rhodin2016general}, or NeRF based~\cite{su2021nerf,su2022danbo,weng2022humannerf} models.
They either explicitly learn a mapping~\cite{su2021nerf,weng2022humannerf,yu2023monohuman} or implicitly use root finding~\cite{li2022tava,wang2022arah} to enable a warp field from observation to canonical space.
Constraining this deformation is difficult as small changes in pose from one frame to the next can, for instance, form high frequency wrinkles on the clothing. 
These details are typically lost or distorted with artifacts.

\begin{figure*}[ht]
    \centering
    \includegraphics[width=0.9\linewidth,trim={0 7.5cm 12cm 0cm},clip]{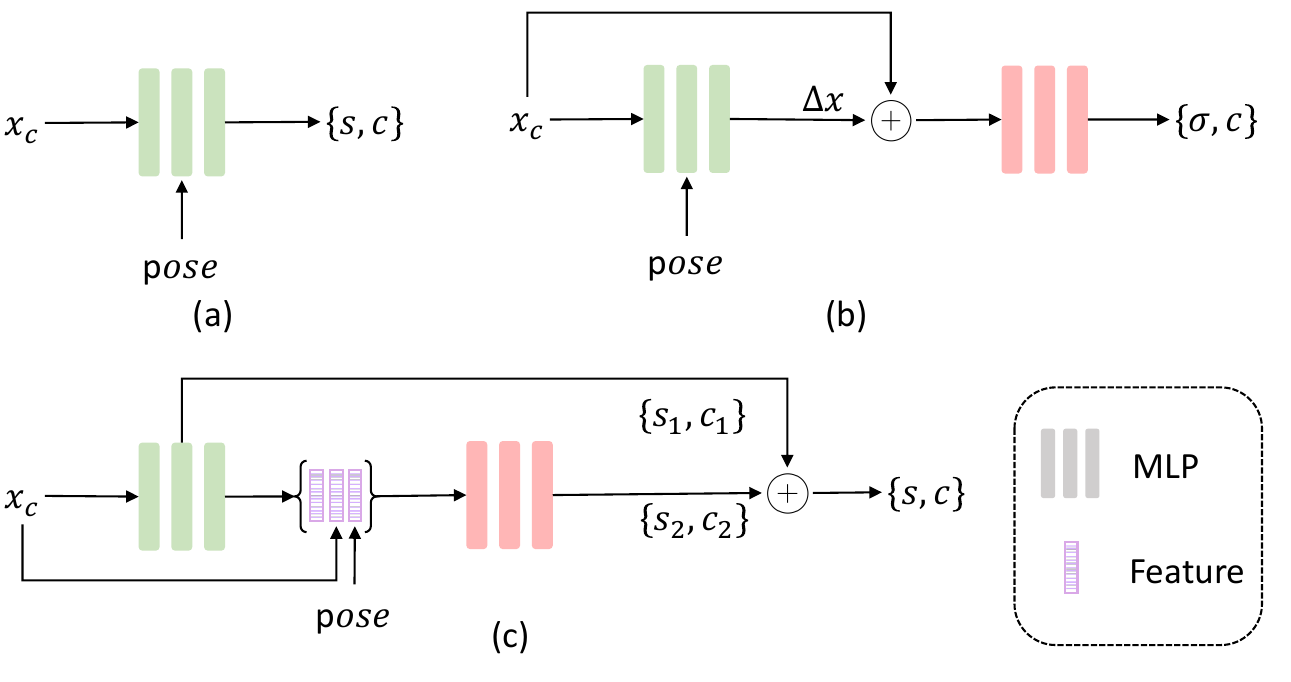}
    \vspace{-0.5em}
    \caption{
    \cj{
    \textbf{Conceptual differences.} Taken a position $x_c$ in canonical space and conditioned on a pose, Vid2Avatar~\cite{guo2023vid2avatar} directly regresses the SDF and appearance values with a uniform frequency band and thus models pose-independent information implicitly (a). 
    In (b), HumanNeRF~\cite{weng2022humannerf} and MonoHuman~\cite{yu2023monohuman} perform decomposition in \textbf{coordinate space} and use a low-frequency network to regress pose-dependent position offset (green) and a high-frequency network (red) for learning pose-independent canonical representations.
    In comparison, we associate the pose-independent information with low frequencies (green) and pose-dependent counterparts with high-frequencies (red) in \textbf{output space} to preserve multi-scale signals (c).
    Here $x_c$ is computed by a skeletal deformation~\cite{weng2022humannerf,guo2023vid2avatar}.
    }
    }
    \vspace{-0.8em}
    \label{fig:overview}
\end{figure*}

Fig.~\ref{fig:teaser} gives an overview of our method.
\cj{
Our goal is to preserve consistent body shapes along with finely matched high-frequency details for monocular videos.
To achieve this, previous methods attempt to either find a mapping from the input positions and poses to geometry and appearance outputs~\cite{guo2023vid2avatar,su2023npc} (sketch in Fig.~\ref{fig:overview} (a)), or explicitly estimate both pose-independent and pose-dependent deformation in coordinate spaces~\cite{weng2022humannerf} (sketch in Fig.~\ref{fig:overview} (b)).
While this, in principle, is a good idea, it cannot maintain frame consistency and adaptive fine-grained details simultaneously.
Our model, in contrast, calculates the pose-independent and pose-dependent components in output spaces, which is inspired by the works on private-shared component analysis~\cite{salzmann2010factorized,bousmalis2016domain}.
}
The pose-independent part is computed from a skeletal deformation~\cite{weng2022humannerf,guo2023vid2avatar} to compute a canonical coordinate and a canonical representation.
Then we explicitly impose pose-independent constraints in output space to improve the frame consistency.  
\cj{
We further condition pose-dependent outputs on pose-independent intermediate features to synthesize more matched non-rigid variations, thus better consistency.
}

\cj{
As shown in Fig.~\ref{fig:mocap-lf_hf}, increasing frequencies in the pose-independent output facilitates more geometric details.
However, it prevents the full model from synthesizing sharper pose-dependent wrinkles as the high frequency leaves the shared representations vulnerable to contamination by 
specific pose-dependent patterns~\cite{salzmann2010factorized,bousmalis2016domain}, making the model harder to generalize.
We thus enforce the per-frame high-frequency details to only be modeled by pose-dependent output factors.
Fig.~\ref{fig:teaser} (a) illustrates the effect of this frequency assumption.
The pose-independent and pose-dependent components characterize the smooth shared shape contours and adaptive high-frequency residuals respectively.
The final result is computed by mixing these two components.
}

To achieve the aforementioned properties, we design a dual-branch network for the pose-independent and pose-dependent outputs respectively.
We follow~\cite{yariv2021volume,guo2023vid2avatar} to enable detailed geometry modeling by applying the Signed Distance Function (SDF) for NeRF-based volume rendering.
We also design an objective for the pose-independent branch to encourage as much information as possible to be encoded in the shared outputs for improved generalization.
Our contributions can be summarized as:
\begin{itemize}%
\item Introducing a novel neural network with two branches, tailored to generate high-fidelity human representations via \cj{the frequency-aware factorized field}.
\item Designing a simple common loss function for the pose-independent branch that maximizes the shared information to improve generality.
\item Demonstrating significant improvement on \cj{the NeRF-based} state-of-the-art methods in novel view synthesis, novel pose rendering and shape reconstruction.
\end{itemize}

\section{Related Work}
\label{Sec:related_work}

In the two following sections, we review related methods in neural field modeling~\cite{nfsurvey} and delve into the latest and most relevant approaches in neural avatar modeling.

\paragraph{Neural Fields.}
Due to the impressive performance, neural fields~\cite{park2019deepsdf,wu2019sagnet,peleg2019net,mescheder2019occupancy,genova2019learning,genova2020local,takikawa2021neural,muller2022instant} are extensively studied to enhance their generalization~\cite{yu2021pixelnerf}, compactness~\cite{wu2022neural,rho2023masked}, level of detail~\cite{takikawa2021neural,muller2022instant, takikawa2022variable}, camera self-calibration~\cite{lin2021barf,yen2021inerf}, and resource efficiency \cite{muller2022instant,SunSC22}.
\cj{
A notable advancement in neural fields is Neural Radiance Fields (NeRF)\cite{mildenhall2020nerf}, developed for rendering images from arbitrary camera views in static scenes. 
Subsequent efforts have extended NeRF to dynamic scenes~\cite{gao2021dynamic,li2022neural,li2021neural, du2021neural,park2021nerfies,park2021hypernerf, tretschk2021non,yuan2021star}, although they do not address significant time-dependent non-rigid deformations commonly encountered in learning human avatar representations~\cite{su2021nerf,su2022danbo,li2022tava, ost2021neural,peng2021neural,xu2021h}.
Additionally, some works attempt to apply the Signed Distance Function (SDF) for NeRF-based models to extract accurate 3D shapes~\cite{yariv2021volume,wang2021neus,wang2022hf,guo2023vid2avatar}.
}

\paragraph{Neural Fields for Avatar Modeling.}
\cj{
In textured avatar modeling, the parametric SMPL body model serves as a common foundation~\cite{zheng2022structured,zheng2023avatarrex}. 
Conversely, approaches such as A-NeRF\cite{su2021nerf} and NARF~\cite{noguchi2021neural} lack a surface prior, directly transforming input query points into relative coordinates of skeletal joints. 
\cj{TAVA~\cite{li2022tava} and ARAH~\cite{wang2022arah} use root finding to enable a warp field from observation to canonical space.}
DANBO~\cite{su2022danbo} models the per-part feature space using a graph neural network for better scalability. 
Later, PM-Avatar~\cite{song2024pose} proposes to modulate query points' frequency transformation based on per-frame pose contexts.
Other methods~\cite{liu2021neural,peng2021neural,peng2021animatable,dong2022totalselfscan} aim to improve results with an image-to-image translation network and a per-frame latent code.
For the human modeling from monocular videos,~\cite{weng2022humannerf,yu2023monohuman,guo2023vid2avatar} learn a canonical space for all frames to improve frame consistencies and testing generalization.
Very recently, some works~\cite{qian20233dgsavatar,kocabas2023hugs,moreau2024human,lei2023gart} apply the Gaussian Splatting framework for better inference speed.
Differing from these methods, we separate a rendering image into a pose-independent factor and pose-dependent counterpart to improve frame consistency and synthesize adaptive details.
See Sec.~\ref{sec:s_field} for more discussions.
}

\section{Method}
\label{Sec:method}

\begin{figure*}[t!]
    \centering
    \includegraphics[width=1.\linewidth,trim={0 5cm 0cm 0cm},clip]{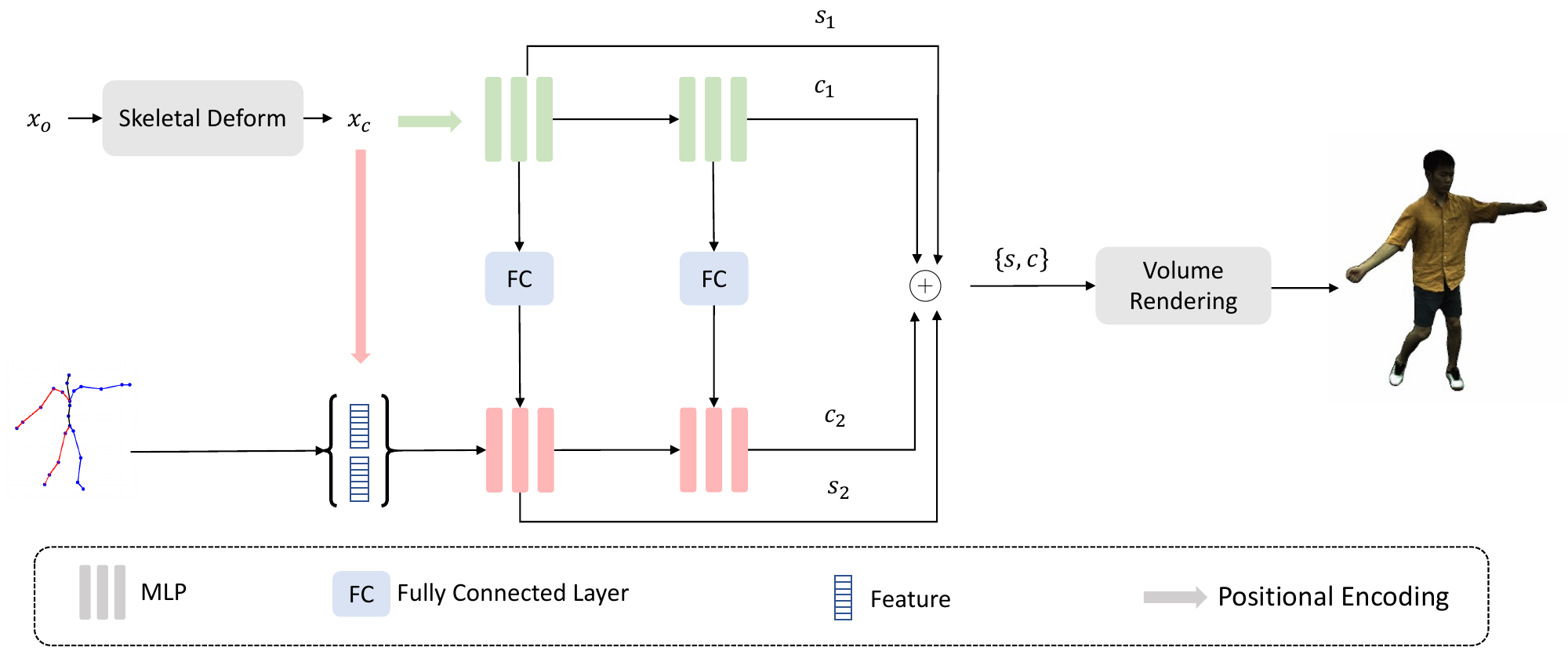}
    \caption{
    \textbf{Architecture overview}. 
    We compute the canonical coordinate $x_c$ of the query point $x_o$ in observation space by performing the skeletal deformation. 
    Then $x_c$ is fed into two branches with
    the low-frequency (green) and high-frequency (red) positional encoding for pose-independent ($\{s_1, c_1\}$) and pose-dependent ($\{s_2, c_2\}$) outputs respectively.
    We input their combinations $\{s, c\}$ to volume rendering to generate images under different view directions and human poses.
    }
    \label{fig:architecture}
\end{figure*}

We aim to reconstruct a 3D animatable avatar by leveraging a collection of $N$ images.
Fig.~\ref{fig:architecture} provides a method overview with three main components. 
First, we estimate the body pose for an input frame, which is represented as the sequence of joint angles $\left[\theta_k\right]_{k=1}^N$.
These joint angles are then used to perform the skeletal deformation for one query point $x_o$ in observation space and obtain the coordinate $x_c$ in canonical space.
Next the computed position $x_c$ is inputted to the two-branch network to output the pose-independent SDF and color value and the pose-dependent counterparts.
Finally, we merge the pose-independent and pose-dependent outputs, which are then mapped to the corresponding density and radiance at that location as in the SDF-based NeRF framework.

\subsection{Skeletal Deformation}
\label{sec:s_deformation}

Modeling 3D avatars in a canonical space is crucial to form a temporally consistent representation.
We follow Vid2Avatar~\cite{guo2023vid2avatar} to perform the skeletal transformation from $x_o$ in observation to $x_c$ in canonical space. Given the $N_B$ joint angles $\theta = [\bf{B}_1, \ldots, \bf{B}_{N_B}]$ of the given body pose, 
the inverse of linear-blend skinning computes
\begin{equation}
x_c = \left ( \sum_{i=1}^{N_B} \omega_i \bf{B}_i \right )^{-1} x_o,
\end{equation}
where $\{\omega_i\}$ denotes the skinning weight of the $i$-th bone and is based on the point-to-point distances to the nearest SMPL vertices in observation space; see~\cite{guo2023vid2avatar} for details.
Compared to the learnable skinning weights~\cite{weng2022humannerf}, this SMPL-based weights can significantly reduce GPU memory consumption. 
This procedure also stabilizes the network training and enables faster convergence speed.

\subsection{Factorized Neural Fields}
\label{sec:s_field}

Our main contribution is to factorize the rendering results of animatable avatars into pose-independent and pose-dependent parts and associate these components with low and high frequencies, respectively.
To achieve this, we design a novel two-branch network as shown in Fig.~\ref{fig:architecture}.
Taken the canonical position $x_c$ as input, we first feed $x_c$ into a low-frequency positional encoding~\cite{mildenhall2020nerf} 
\begin{equation}
\bar{\gamma} (x_c) = (x_c, \sin (2^0 \pi x_c), \cos(2^0 \pi x_c), \dots, \sin (2^{L_{ind}-1} \pi x_c), \cos (2^{L_{ind}-1} \pi x_c)),
\end{equation}
where $L_{ind}$ indicates the highest mapping frequency in $\bar{\gamma} (x_c)$.
Then the processed feature $\bar{\gamma} (x_c)$ is fed to the upper branch $\overline{\mathrm{MLP}}$, which is an MLP with ReLU activations, to estimate the pose-independent SDF value $\{s_1\}$ and RGB value $\{c_1\}$.
To leverage the relationships between pose-independent and pose-dependent information and synthesize adaptive details, the upper branch also learns to output a feature vector for the SDF network and color network on the bottom branch respectively
\begin{equation}
s_1, \bar{f}_{sdf} = \overline{\mathrm{MLP}}_{sdf} (\bar{\gamma} (x_c)),
\quad
c_1, \bar{f}_{c} = \overline{\mathrm{MLP}}_{c} (\bar{f}_{sdf}, \overrightarrow{n}_1).
\end{equation}
Here $\overrightarrow{n}_1$ indicates the normalized gradient of $s_1$ computed at $x_c$. 
$\overline{\mathrm{MLP}}_{sdf}$ and $\overline{\mathrm{MLP}}_{c}$ denote the SDF network and color network of the upper branch, respectively.

As mentioned above, we observe that the pose-dependent patterns, such as dynamically changing wrinkles on clothes, should process higher frequencies than its pose-independent analogue.
Thus, we feed $x_c$ into another high-frequency positional encoding as:
\begin{equation}
\tilde{\gamma} (x_c) = (x_c, \sin (2^0 \pi x_c), \cos(2^0 \pi x_c), \dots, \sin (2^{L_{d}-1} \pi x_c), \cos (2^{L_{d}-1} \pi x_c)).
\end{equation}
Similar to $L_{ind}$ in $\bar{\gamma} (x_c)$, $L_{d}$ here stands for the highest transformation frequency of $\tilde{\gamma} (x_c)$. We then give the concatenation of $\tilde{\gamma} (x_c)$, $\bar{f}_{sdf}$ and the input pose $\theta$ to the bottom branch $\widetilde{\mathrm{MLP}}$ to output pose-dependent SDF value $s_2$ and color $c_2$ as
\begin{equation}
s_2, \tilde{f}_{sdf} = \widetilde{\mathrm{MLP}}_{sdf} ([\tilde{\gamma} (x_c), \bar{f}_{sdf}, \theta],
\quad
c_2 = \widetilde{\mathrm{MLP}}_{c} ([\tilde{f}_{sdf}, \bar{f}_c, \overrightarrow{n}, \theta].
\end{equation}

To combine the low-frequency pose-independent and high-frequency pose-dependent outputs, we compute the final SDF and color outputs for further processing as
\begin{equation}
s = s_1 + s_2, \quad c = c_1 + c_2.
\end{equation}
Note that we apply the gradient vector $\overrightarrow{n}$ of the learned signed distance function $s$ to the color network to facilitate the disentanglement of geometry and appearance~\cite{yariv2021volume}.
Following HumanNeRF, we set $L_{d}=10$ and halve $L_{ind}=5$ for simplicity across all experiments.

\begin{figure*}[t]
    \centering
    \includegraphics[width=0.9\linewidth,trim={0 26cm 10cm 0cm},clip]{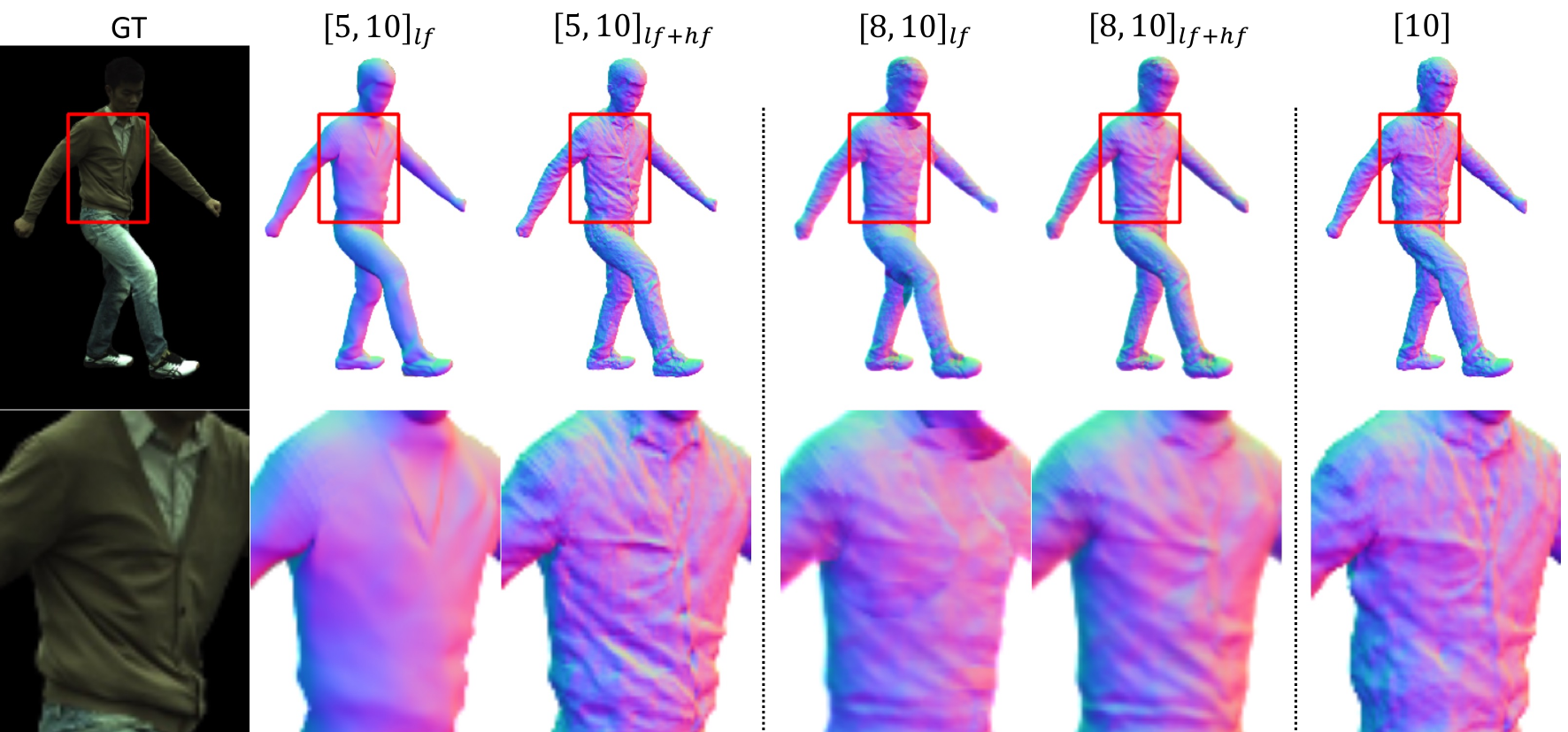}
    \caption{
    \cj{
    \textbf{Frequency constraints.} 
    \cj{To validate our frequency assumption, we train a set of two-branch models with different $L_{ind}$ and $L_{d}$.
    For simplicity, we denote a model with $L_{ind}=x$ and $L_{d}=y$ as $[x, y]$.}
    Adhering to our network design, the pose-independent branch outputs the low-frequency base normal map as $[5, 10]_{lf}$ while our full model estimates an output with all frequencies as $[5, 10]_{lf+hf}$.
    Increasing frequency in the pose-independent output, denoted as $[8, 10]$, can yield more grainy geometric patterns in $[8, 10]_{lf}$ but stops the full model from generating sharp pose-dependent wrinkles in $[8, 10]_{lf+hf}$.
    Simply training the pose-dependent branch ($[10]$ with $L_{d}=10$) fails to synthesize desired multi-scale patterns.
    See Sec.~\ref{sec:s_field} and Sec.~\ref{sec:supp_abtest} in appendix for more discussions.
    }
    \vspace{-0.5em}
    }
    \label{fig:mocap-lf_hf}
\end{figure*}

\paragraph{Relation to baselines.}
Fig.~\ref{fig:overview} illustrates the conceptual differences between our method and two representative baselines. 
\cj{
Both Vid2Avatar and our method apply SDF to possess constant shape contours and improve frame consistencies.
However, Vid2Avatar directly maps the input positions and poses to geometry and appearance with uniform frequencies and does not model the pose-invariant information for better generalization.
Thus it either introduces unwanted artifacts or blurs the desired high-frequency textures (e.g. $2^{nd}$ row in  Fig.~\ref{fig:mocap}). 
The density-based HumanNeRF decomposes the deformation into a rigid pose-independent part and its non-rigid pose-dependent counterpart in coordinate space.
While it can synthesize fine-grained details, outputs from the highly non-linear MLP twist the body shape with fuzzy boundaries as shown in the $1^{st}$ row of Fig.~\ref{fig:mocap}.
In contrast, our method has superior frame consistency in body outlines and better generalization to precise textures due to explicit pose-independent modeling and accurate frequency associations in output space.
Additionally, pose-dependent outputs based on intermediate pose-independent features also facilitate per-frame adaptive patterns over baselines.
}

\paragraph{Discussions on frequency bands.} 
\cj{
We enforce the high-frequency information to only be encoded in pose-dependent output for improving the synthesis of pose adaptive results.
\hr{A few high-frequency} patterns, such as the facial structures and shoe textures in Fig.~\ref{fig:mocap-lf_hf}, only deform lightly, and \hr{could} be encoded into \hr{a}
pose-independent output with high frequency.
However, we only use one single positional encoding \hr{setting} for all parts
\hr{and such part-dependent properties are not known as prior and may change from part to part.} \hr{We} trade off how much information each part contributes to the shared features.
As shown in Fig.~\ref{fig:mocap-lf_hf}, simply increasing frequency helps create more realistic facial patterns and introduce more geometric details in the highlighted torso.
But it also goes against the generalization to novel poses and leads to blunt pose-dependent wrinkles as in $[8, 10]_{lf+hf}$.
Similar to Vid2Avatar, the one-branch network with uniform frequency band fails to reproduce multi-scale geometries.
Moreover, the distinct frequency assignments help penalize redundant information between the pose-independent and pose-dependent branches.
See the appendix for additional results.

}

\subsection{SDF-based volume rendering}
\label{sec:sdf_volume}

With the output SDF and color signals $\{s, c\}$, we first compute the density via a learnable transformation~\cite{yariv2021volume}, namely
\begin{equation}
\sigma = \frac{1}{\beta} \cdot \Psi_\beta (-s),
\end{equation}
where $\beta$ is a learnable parameter and $\Psi$ is the Cumulative Distribution Function (CDF) of the Laplace distribution with zero mean and $\beta$ scale.

Following the existing neural radiance rendering pipelines for human avatars~\cite{su2021nerf,su2022danbo,wang2022arah,song2024pose}, we output the image of a subject for a ray $r$ as in the original NeRF:
\begin{equation}
\hat{C} \left ( r \right ) = \sum_{i=1}^{n} \mathcal{T}_i \left ( 1 - \mathrm{exp} (-\sigma_i \delta _i) \right ) \bc_i,
\mathcal{T}_i = \mathrm{exp}(-\sum_{j=1}^{i-1} \sigma_j \delta_j).
\end{equation}
Here, $\hat{C}$ and $\delta_i$ indicate the synthesized image and the distance between adjacent samples along a given ray respectively.
Finally, we compute the $L_1$ loss $\left \| \cdot \right \|_1$ for training as
\begin{equation}
\mathcal{L}_{\mathrm{rec}} = \sum_{\ray \in \Rays } \left \| \hat{\Color}(r) - \Color_{gt}(r) \right \|_1,
\end{equation}
where $\Rays$ is the whole ray set and $\Color_{gt}$ is the ground truth.

Following~\cite{yariv2021volume,guo2023vid2avatar}, we apply the Eikonal loss to satisfy the Eikonal equation such that the learned $s=s_1 + s_2$ can approximate a signed distance function as
\begin{equation}
\mathcal{L}_{\mathrm{eik}} = \mathbb{E}_{x_c} (\left \| \frac{\mathrm{d} s}{\mathrm{d} x_c} \right \| - 1)^2.
\end{equation}

We observe that it is crucial to maximize the amount of information in the pose-independent (upper) branch to improve generality.
To avoid that the upper branch degrades and the pose-dependent branch takes most information, we render the human image by volume rendering with the pose-independent SDF and color components $\{s_1, c_1\}$ and compute the $L_1$ loss for common data as
\begin{equation}
\mathcal{L}_{\mathrm{com}} = \sum_{\ray \in \Rays } \left \| \bar{\Color}(r) - \Color_{gt}(r) \right \|_1,
\end{equation}
where $\bar{\Color}$ is the synthesized image created by $\{s_1, c_1\}$.

Besides the aforementioned $L_1$ loss, a perceptual loss, LPIPS~\cite{zhang2018unreasonable}, is employed to provide robustness to slight misalignments and shading variation and to improve detail in the reconstruction as $\mathcal{L}_{\mathrm{LPIPS}}$.

Thus, given a video sequence of a human, we aim to optimize the following combined loss function:
\begin{equation}
\mathcal{L} = \mathcal{L}_{\mathrm{rec}} + \lambda_{\mathrm{eik}} \mathcal{L}_{\mathrm{eik}} +
\lambda_{\mathrm{com}} \mathcal{L}_{\mathrm{com}} +
\lambda_{\mathrm{LPIPS}} \mathcal{L}_{\mathrm{LPIPS}}.
\end{equation}
Here $\lambda_{\mathrm{eik}}$, $\lambda_{\mathrm{com}}$ and $\lambda_{\mathrm{LPIPS}}$ are weights for Eikonal loss, common loss and LPIPS loss, respectively.

\section{Results}
\label{Sec:results}

\cj{
In this section, we compare our approach with several state-of-the-art methods, including HumanNeRF~\cite{weng2022humannerf}, MonoHuman~\cite{yu2023monohuman}, NPC~\cite{su2023npc}, Vid2Avatar~\cite{guo2023vid2avatar} and PM-Avatar~\cite{song2024pose}, for rendering results and 3D shape reconstruction.
We also conduct ablation studies to 
analyze and discuss the effects of factorized avatar representation, common loss function and the dependencies between pose-independent and pose-dependent branches.
Source code will be released with the publication.
}

\subsection{Experimental Settings}
\label{sec:results_settings}

We assess the effectiveness of our approach using well-established benchmarks for body modeling. Following HumanNeRF and MonoHuman, we conduct evaluations across the eight sequences of the ZJU-Mocap dataset~\cite{peng2021neural}. 
Additionally, we utilize two publicly available YouTube sequences as an in-the-wild dataset to gauge performance on everyday monocular videos. 
We follow Vid2Avatar to process all sequences and obtain approximate camera and body pose with off-the-shelf estimators. 
We employ the SAM model~\cite{kirillov2023segment} to generate foreground maps of all images for precise segmentation.
\begin{wraptable}{r}{0.5\linewidth}
\vspace{-0.1in}
\caption{\textbf{Novel-view and novel-pose synthesis results, averaged over the ZJU-Mocap test set~\cite{peng2021neural}.} 
Our factorized fields enable better results with over 10\% improvement in LPIPS and KID than the best baseline for both novel view synthesis and novel pose rendering.}
\centering
\resizebox{\linewidth}{!}{
\setlength{\tabcolsep}{3pt}
\begin{tabular}{lcccccccc}
\toprule
& \multicolumn{4}{c}{Novel view} 
& \multicolumn{4}{c}{Novel pose} \\
\cmidrule(lr){2-5}  \cmidrule(lr){6-9}%
  & PSNR$\uparrow$  & SSIM$\uparrow$  & LPIPS$\downarrow$ & KID$\downarrow$ & PSNR$\uparrow$  & SSIM$\uparrow$  & LPIPS$\downarrow$ & KID$\downarrow$\\
\midrule
HumanNeRF& 29.94 & 0.967 & 31.81 & 14.23 & 29.45 & 0.966 & 32.18 & 12.32 \\
\rowcolor{gray}
MonoHuman& 30.03 & 0.967 & 33.47 & 13.18 & 29.73 & 0.967 & 34.25 & 12.61\\
NPC& 30.01 & 0.967 & 37.18 & 53.24 & 29.61 & 0.967 & 36.52 & 49.79 \\
\rowcolor{gray}
PM-Avatar& 30.27 & 0.969 & 38.38 & 39.64 & 29.87 & 0.969 & 39.26 & 40.16 \\
Vid2Avatar& 29.76 & 0.969 & 35.61 & 27.65 & 29.53 & 0.969 & 35.69 & 31.51\\
\midrule
Ours& \textbf{30.11} & \textbf{0.970} & \textbf{29.64} & \textbf{11.72} & \textbf{29.98} & \textbf{0.970} & \textbf{28.60} & \textbf{11.09}\\
\bottomrule
\end{tabular}
\label{tab:mocap-score}
}
\vspace{-0.2in}
\end{wraptable}

To ensure comparison fairness, we follow former experimental settings, including dataset splits and metrics~\cite{weng2022humannerf,yu2023monohuman,su2023npc}. Our evaluation covers standard image metrics like pixel-wise Peak Signal-to-Noise Ratio (PSNR) and Structural Similarity Index Metric (SSIM)\cite{wang2004image} to assess image quality. 
We also utilize perceptual metrics such as Learned Perceptual Image Patch Similarity (LPIPS)\cite{zhang2018unreasonable}, Kernel Inception Distance (KID)\cite{bińkowski2018demystifying}, and Fréchet Inception Distance (FID)\cite{heusel2017gans} to evaluate structural accuracy and textured details of generated images. 
All metrics are computed across entire generated images. And both LPIPS and KID metrics are multiplied by 1000 for better comparisons.
\cj{
Similar to ARAH~\cite{wang2022arah}, we additionally report Chamfer Distance (CD) and Normal Consistency (NC) to estimate the quality of reconstructed shapes.
}

\begin{figure*}[t!]
    \centering
    \includegraphics[width=0.9\linewidth,trim={0 23.5cm 0cm 0cm},clip]{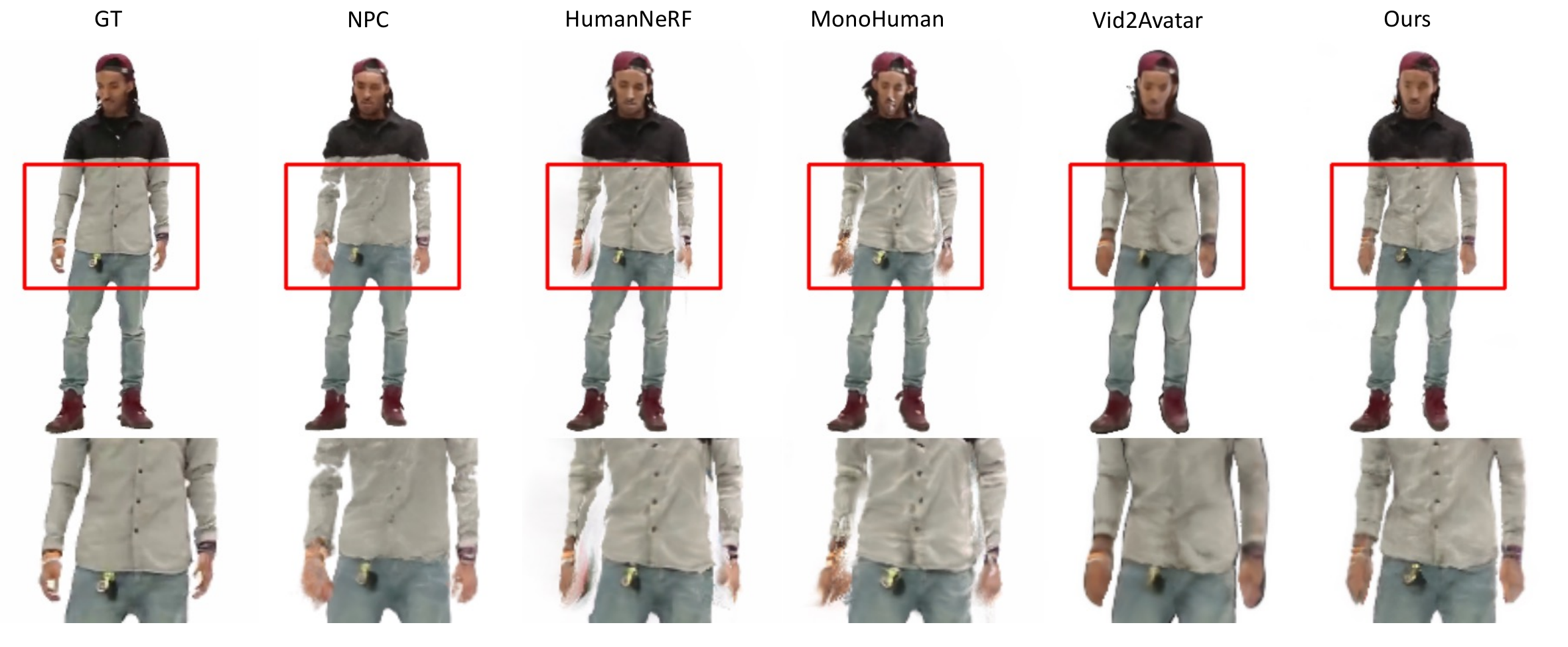}
    \caption{
    \textbf{Novel pose rendering on Youtube sequences.}
    While baselines distort the marked arms with floating noise, our method yields more visually appealing body outlines.
    We also improve Vid2Avatar with more realistic textures like cloth buttons.
    }
    \vspace{-0.5em}
    \label{fig:youtube-npr}
\end{figure*}

\subsection{Novel View Synthesis}
\label{sec:results_NVS}

We utilize ZJU-Mocap sequences as a multi-view dataset to evaluate the generalization capability under different camera views.
Specifically, we use images captured by the first provided camera (“camera 1”) for training and the remaining images for evaluation.

\begin{figure*}[t]
    \centering
    \includegraphics[width=0.9\linewidth,trim={0 24.2cm 21.4cm 0cm},clip]{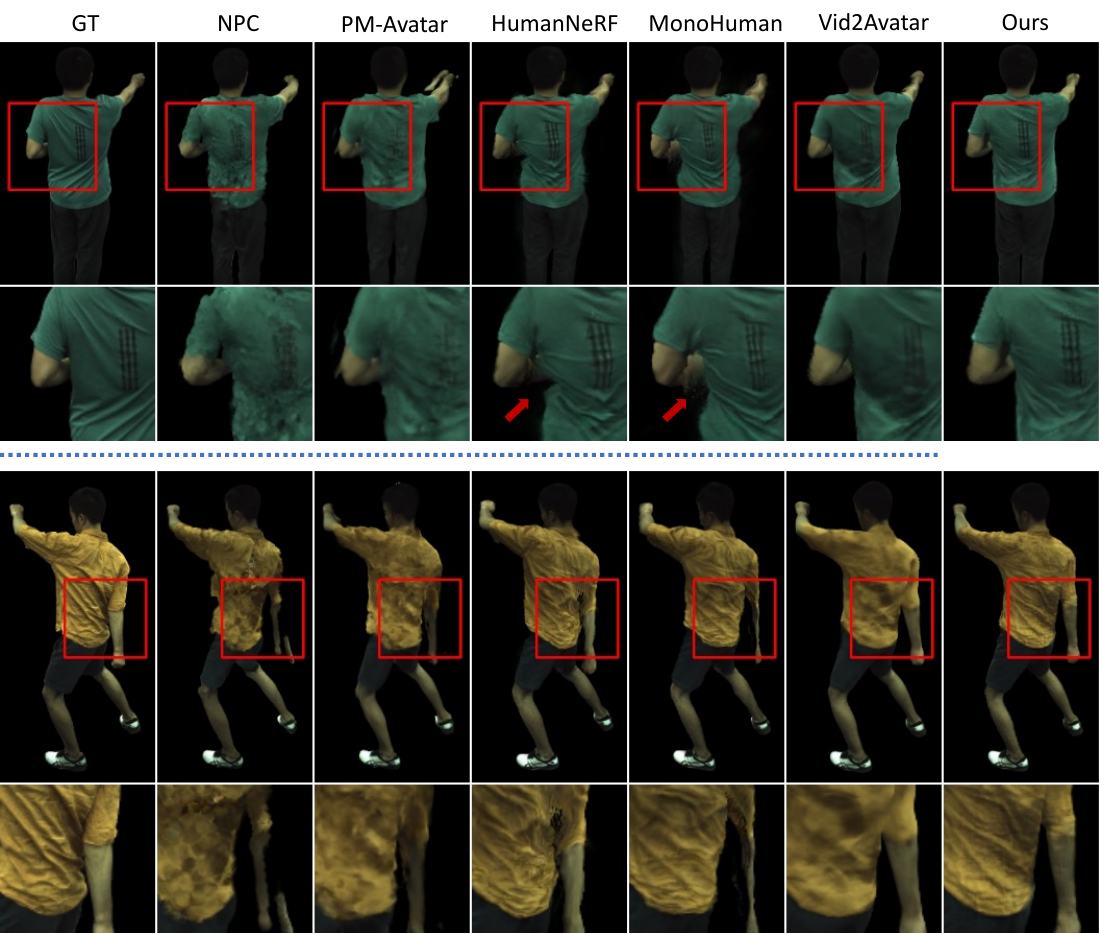}
    \caption{
    \textbf{Visual comparisons on ZJU-Mocap.} 
    Our method can render superior sharp contours and synthesize more adaptive textures under novel camera views (upper) and novel avatar poses (bottom).
    }
    \vspace{-1.5em}
    \label{fig:mocap}
\end{figure*}

We visualize the results in the upper row of Fig.~\ref{fig:mocap}. 
Comparing to baselines, our method shows superior capabilities in recovering fine-grained details (e.g. the vertical patterns). 
Additionally, our method better preserves the body shape, such as cloth contours.
We attribute this to the explicit separation in output space which mitigates grainy artifacts and reserves consistent large-scale outlines.
We compare quantitatively in Tab.~\ref{tab:mocap-score} to further support our previous findings.

\begin{table}[b]
\vspace{-1.5em}
\caption{
\textbf{(a)} The first two columns list the unseen pose results on the Youtube videos.
The third reports the $L_2$ Chamfer Distance (CD) and Normal Consistency (NC) over ZJU-Mocap sequences for geometry reconstruction evaluation.
Our model shows better overall perceptual quality and shape reconstruction from monocular videos. 
$^*$The imperfect pseudo-ground-truth smooths details, benefiting Vid2Avatar for the NC metric; see the appendix.
\textbf{(b)} Ablation study on ZJU-Mocap sequence.
Our full model outperforms all ablated baselines across all metrics.
}
\centering
\resizebox{1.0\linewidth}{!}{
\setlength{\tabcolsep}{3pt}
\begin{tabular}[b]{ccccccc}
\toprule
& \multicolumn{2}{c}{Story} & \multicolumn{2}{c}{Invisible} & \multicolumn{2}{c}{Geometry} \\
\cmidrule(lr){2-3}\cmidrule(lr){4-5}\cmidrule(lr){6-7}%
 & LPIPS$\downarrow$  & FID$\downarrow$  & LPIPS$\downarrow$  & FID$\downarrow$  & CD$\downarrow$  & NC$\uparrow$ \\
\midrule
HumanNeRF & 31.35 & 63.28 & 33.72 & 72.29 & 0.242 & 0.649 \\
\rowcolor{gray}
MonoHuman & 32.73 & 65.23 & 34.39 & 79.94 & 0.318 & 0.636 \\
NPC & 29.59 & 53.62 & 35.28 & 80.17 & 0.079 & 0.795 \\
\rowcolor{gray}
Vid2Avatar & 36.85 & 187.24 & 40.52 & 198.51 & 0.053 & \textbf{0.878}$^*$ \\
\midrule
\textbf{Ours}& \textbf{28.52}& \textbf{56.57} & \textbf{31.74}& \textbf{69.13}& \textbf{0.047}& 0.863\\
\midrule
\multicolumn{7}{c}{(a)}
\end{tabular}
\qquad
\begin{tabular}[b]{ccccccc}
\toprule
& \multicolumn{3}{c}{Novel view} & \multicolumn{3}{c}{Novel pose} \\
\cmidrule(lr){2-4}\cmidrule(lr){5-7}
 & PSNR$\uparrow$ & SSIM$\uparrow$ & LPIPS$\downarrow$  & PSNR$\uparrow$ & SSIM$\uparrow$ & LPIPS$\downarrow$ \\
\midrule
$\mathrm{w/o\ \{s_1, c_1\}}$ & 30.13 & 0.965 & 31.00 & 29.36 & 0.962 & 31.01 \\
\rowcolor{gray}
$\mathrm{w/o\ \{s_2, c_2\}}$ & 29.56 & 0.963 & 32.46 & 29.26 & 0.962 & 31.30 \\
$\mathrm{w/o\ \mathcal{L}_{\mathrm{com}}}$ & 30.08 & 0.964 & 31.40 & 29.47 & 0.963 & 30.90 \\
\rowcolor{gray}
$\mathrm{w/o\ feat}$ & 29.49 & 0.962 & 33.41 & 29.15 & 0.961 & 32.77 \\
$\mathrm{Pose}_{lf}$ & 30.11 & 0.965 & 29.83 & 29.28 & 0.963 & 30.53  \\
\midrule
\textbf{Ours}& \textbf{30.68}& \textbf{0.967} & \textbf{29.74}& \textbf{29.69}& \textbf{0.965}& \textbf{29.41}\\
\midrule
\multicolumn{7}{c}{(b)}
\end{tabular}%
\label{tab:ablation_shape_youtube}%
}%
\end{table}%

\subsection{Novel Pose Rendering}
\label{sec:results_NPR}

Rendering under novel poses is critical to many down-streamed applications like computer animations.
To evaluate the generalization to unseen human poses, we train all models on the first part of a video and test on the remaining frames.
During evaluation, only the 3D human pose is used as input.

\cj{
Fig.~\ref{fig:mocap} illustrates the visual comparisons for the ZJU-Mocap dataset, where our method shows more desirable results in terms of sharp body boundaries (e.g. arms) and fine-grained textures (e.g. wrinkles). 
}
In contrast, baselines either introduce blurry patterns (e.g. Vid2Avatar) or distort the shape contours with noisy artifacts (e.g. MonoHuman).
Fig.~\ref{fig:youtube-npr} additionally presents the results of Youtube sequences.
While our method succeeds in generating reasonable multi-scale patterns, baselines severely distort the highlighted arms under challenging poses.
Tab.~\ref{tab:mocap-score} and Tab.~\ref{tab:ablation_shape_youtube} (a) further quantitatively verify our strong generalization to unseen poses.
Note that none of the methods achieve perfect alignment with wrinkle locations due to their chaotic formation on unseen poses. Addressing this issue using physics falls outside the scope of this paper.
Notably, generalizing to novel poses is more challenging than novel view synthesis and the case where we attain the largest relative improvements compared to baselines. 
This empirical evidence substantiates the effectiveness of our frequency-aware
factorized avatar representation.

\begin{figure*}[t!]
    \centering
    \includegraphics[width=0.9\linewidth,trim={0 26.5cm 0cm 0cm},clip]{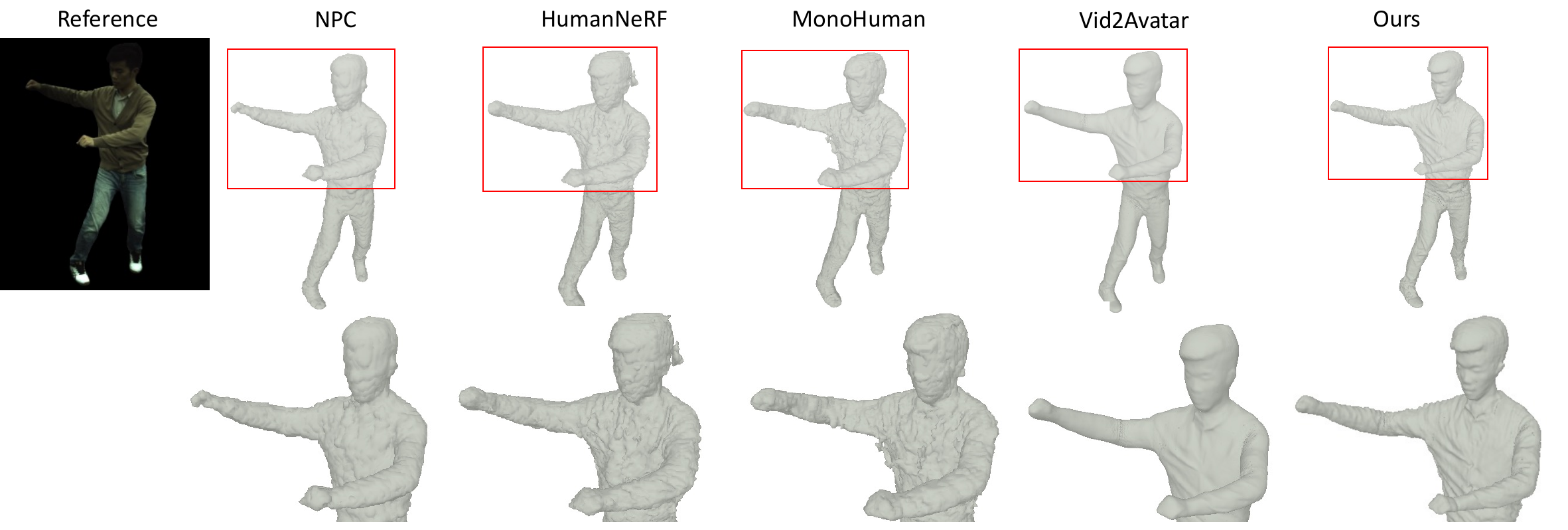}
    \caption{
    \textbf{Comparisons on geometry reconstruction.} 
    Our method yields more precise shape estimates with fine-grained geometric details.
    }
    \vspace{-1.0em}
    \label{fig:mocap-geometry-single}
\end{figure*}

\subsection{Geometry Comparisons}
\label{sec:results_geometry}

In Fig.~\ref{fig:mocap-geometry-single}, we analyze the 3D meshes reconstructed with our approach against reconstructions from the baselines. 
Our method better captures the smooth body surfaces and detailed geometry (e.g. the wrinkles).
In contrast, the baselines predict more noisy blobs near the body surface whose structured patterns (e.g. facial expressions) cannot be faithfully recognized.
While Vid2Avatar does provide a complete body outline, it tends to overly smooth out sharp textures and generate flat patterns.
The `Geometry' column in Tab.~\ref{tab:ablation_shape_youtube} (a) additionally complies with our empirical advantages in a quantitative manner. 
Note that we follow ARAH~\cite{wang2022arah} to compute pseudo ground truth which by itself smooths surface details, more than our method does.
Being consistent with better generalization of novel view synthesis and novel pose rendering, the improvements of geometry reconstruction suggest that more precise modeling of geometry is beneficial for the visual fidelity.

\subsection{Ablation studies}
\label{sec:results_ab_test}

We conduct ablation studies with the following ablated models: \textbf{1.}~Only preserve the bottom branch network with pose-dependent deformations as `$\mathrm{w/o\ \{s_1, c_1\}}$';
\textbf{2.}~Only preserve the upper branch network with pose-independent deformations as `$\mathrm{w/o\ \{s_2, c_2\}}$';
\textbf{3.}~We remove the common loss function as `$\mathrm{w/o\ \mathcal{L}_{\mathrm{com}}}$';
\textbf{4.}~For the bottom branch, we only input the target pose and $x_c$ as `$\mathrm{w/o\ feat}$';
\textbf{5.}~We feed the body pose to the pose-independent branch instead of the pose-dependent branch as $\mathrm{Pose}_{lf}$.
These experiments are performed to evaluate the effectiveness of the factorized fields, the common loss $\mathcal{L}_{\mathrm{com}}$ and the dependency between two branches respectively.
$\mathrm{Pose}_{lf}$ is further trained to evaluate the importance of common information among frames.
We perform comparisons in both novel view synthesis and novel pose rendering on the ZJU-Mocap S394 sequence.
The quantitative results shown in Tab.~\ref{tab:ablation_shape_youtube} (b) consistently highlight the importance of all network components.
We additionally offer more ablation study  results and discussions in the appendix.

\section{Conclusions}
\label{Sec:conclusion}

We introduce a novel two-branch framework to enhance the accuracy of avatar representation learning. 
Our primary contribution lies in a unique 
frequency-aware field factorization design, which enhances frame consistency and boosts the ability to produce adaptive details. 
In comparison to existing methods, our approach demonstrates empirical advantages in novel view synthesis, novel pose rendering, and shape reconstruction.

\clearpage

{\small
\bibliographystyle{unsrt}
\bibliography{main}
}

\clearpage
\appendix
\renewcommand\thefigure{\Alph{figure}}
\setcounter{figure}{0}
\renewcommand\thetable{\Alph{table}}
\setcounter{table}{0}
\setcounter{footnote}{0}

\centering
\Large
\textbf{Representing Animatable Avatar via Factorized Neural Fields} \\
\vspace{0.5em} (Supplementary Material) \\
\vspace{1.0em}

\justifying
\normalsize

\cj{
In the appendix, we first provide additional implementation details for our method.
}
Then more ablation study results are shown to emphasize the significance of the frequency-aware field factorization.
To demonstrate our generalization to camera views and human poses, we show the time consistency by offering rendering results on ZJU-Mocap sequences from different time steps, as well as the novel view synthesis results of Youtube sequences.
Additionally, we present detailed comparisons for 3D shape reconstruction.
Following HumanNeRF~\cite{weng2022humannerf} and Vid2Avatar~\cite{guo2023vid2avatar}, we show the appearance of our reconstructed 3D avatars in canonical space.
Additionally, we reveal the necessity of using masks for training Vid2Avatar models.
Finally, we discuss the limitations and social impacts of this project.
See the attached video for the animation results.

\cj{
\section{Implementation Details}
\label{sec:supp_implementation_details}

We maintain the same hyperparameter settings across various experiments, including the weights of loss functions, the number of training iterations, the network capacity and learning rate.
Our method is implemented using PyTorch~\cite{paszke2019pytorch}. We utilize the Adam optimizer~\cite{kingma2014adam} with default parameters $\beta_1 = 0.9$ and $\beta_2 = 0.99$. 
We employ the step decay schedule to adjust the learning rate, where the initial learning rate is set to $5 \times 10^{-4}$ and we drop the learning rate to $50\%$ in the end of the $200$ and $400$ epochs.
We set $N_B=24$ as~\cite{su2021nerf,su2022danbo}.
Like~\cite{yariv2021volume,guo2023vid2avatar}, the SDF networks in both upper and bottom branches are activated by Softplus while the color networks are activated by ReLU~\cite{agarap2018deep}, 
The initially estimated human poses are refined during training, alongside the network parameters.
We train our network on two NVIDIA Tesla V100 GPUs for 30 hours.
}

\section{Ablation Study Results}
\label{sec:supp_abtest}

To more comprehensively evaluate the effectiveness of our frequency-aware field factorization concept, we provide visual comparisons here besides the reported scores in Tab. 3 (b) of the main text.
Specifically, we group all ablation results into the following four parts to highlight our implementation motivations.

\begin{figure*}[hb]
    \centering
    \includegraphics[width=0.9\linewidth,trim={0 28cm 14.5cm 0cm},clip]{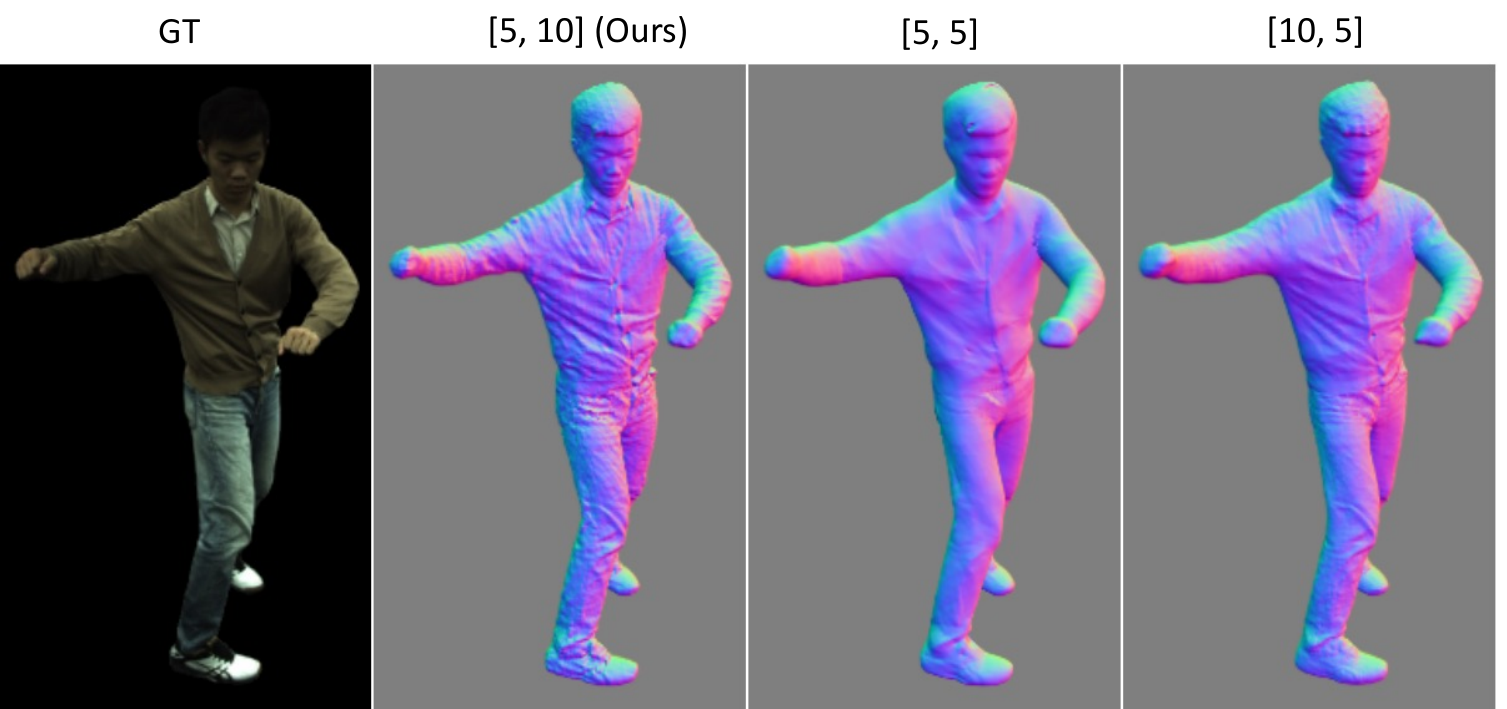}
    \caption{
    \cj{
    \textbf{Ablation study on frequency assignments.}
    Our full model associating pose-independent (pose-dependent) outputs with low (high) frequencies can more faithfully reproduce the target geometric details.
    In contrast, the ablation baselines fail to capture fine-grained wrinkles.
    }
    }
    \label{fig:ab_freq}
\end{figure*}

\begin{figure*}[p]
    \centering
    \includegraphics[width=0.85\linewidth,trim={0 19cm 0cm 0cm},clip]{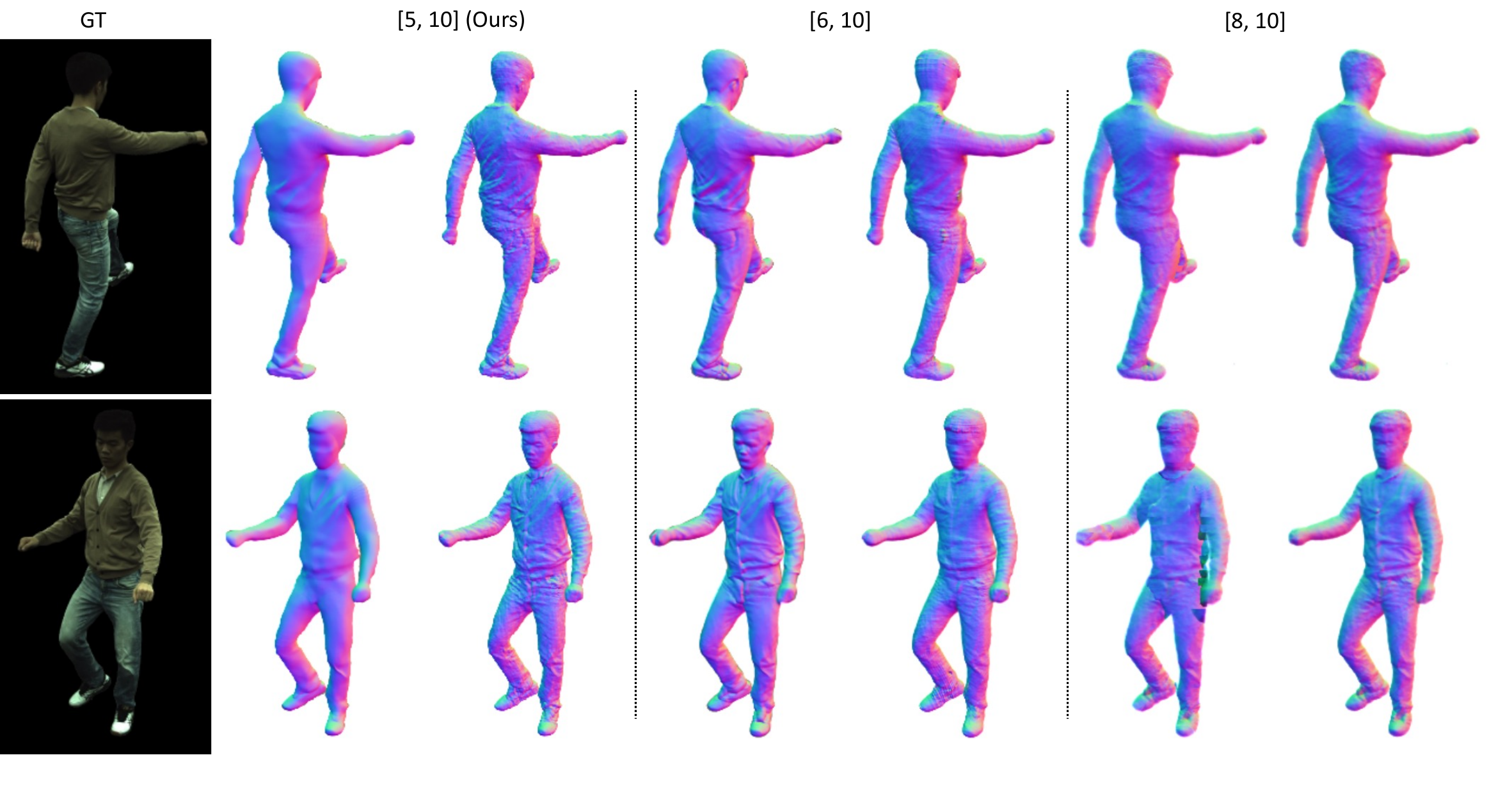}
    \caption{
    \cj{
    \textbf{Frequency assignments for pose-independent outputs.} 
    Increasing the maximum frequency in positional encoding enables the pose-independent output to encode more high-frequency patterns but in turn prevents synthesizing pose-dependent details (e.g. wrinkles on the jacket).
    }
    }
    \label{fig:supp-ab-comm_freq}
\end{figure*}

\cj{
\textbf{1.}~To validate our frequency association design, we train a set of two-branch models with different $L_{ind}$ and $L_{d}$. For simplicity, we denote a model with $L_{ind}=x$ and $L_{d}=y$ as $[x, y]$.
Instead of assigning low (high) frequencies to pose-independent (pose-dependent) deformation outputs, the ablated models fail to reproduce the accurate texture directions in Fig.~\ref{fig:ab_freq} and Fig.~\ref{fig:supp-ab-comm_freq}.
Tab.~\ref{tab:supp_ablation_freq} further quantitatively supports the importance of our frequency assumption.
}

\begin{figure*}[t]
    \centering
    \includegraphics[width=0.9\linewidth,trim={0 19cm 0cm 0cm},clip]{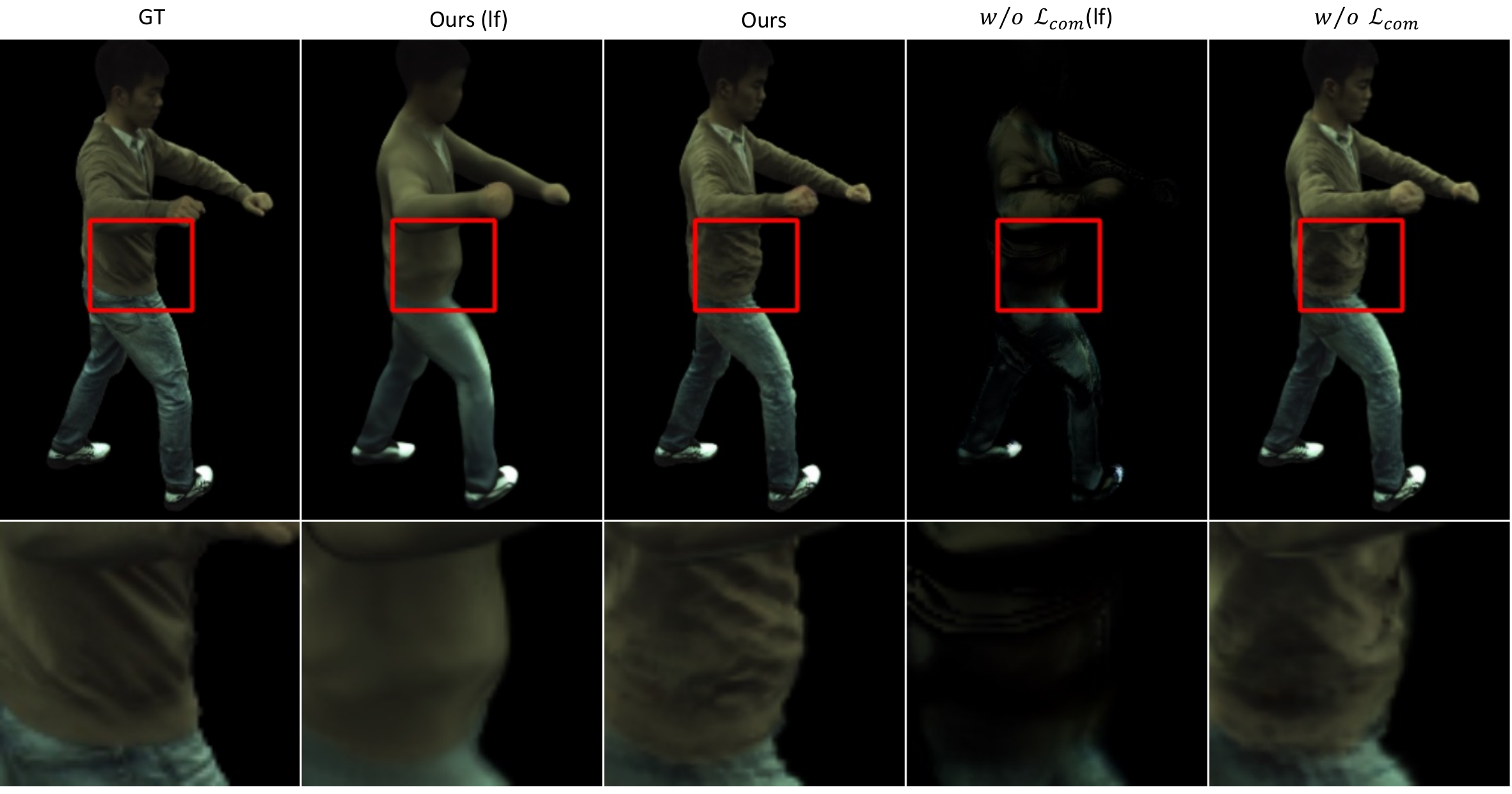}
    \caption{
    \textbf{Ablation study on the Common loss $\mathrm{w/o\ \mathcal{L}_{\mathrm{com}}}$.}
    Our full model with $\mathrm{w/o\ \mathcal{L}_{\mathrm{com}}}$ can effectively preserves semantic textures with precise pattern directions, whereas the ablated model results in notable texture distortions.
    }
    \label{fig:ab_Lcom}
\end{figure*}

\textbf{2.}~In Fig.~\ref{fig:ab_Lcom}, disabling the loss function $\mathcal{L}_{\mathrm{com}}$ for the pose-independent branch results in noticeable distortions with black artifacts (last column) due to network degradation.
Furthermore, we display the RGB outputs from the pose-independent branch of both models.
It is evident that the ablated model without $\mathcal{L}_{\mathrm{com}}$ fails to produce meaningful outputs, whereas our full model maintains consistent shape outlines.

\begin{figure*}[p]
    \centering
    \includegraphics[width=0.85\linewidth,trim={0 32cm 18cm 0cm},clip]{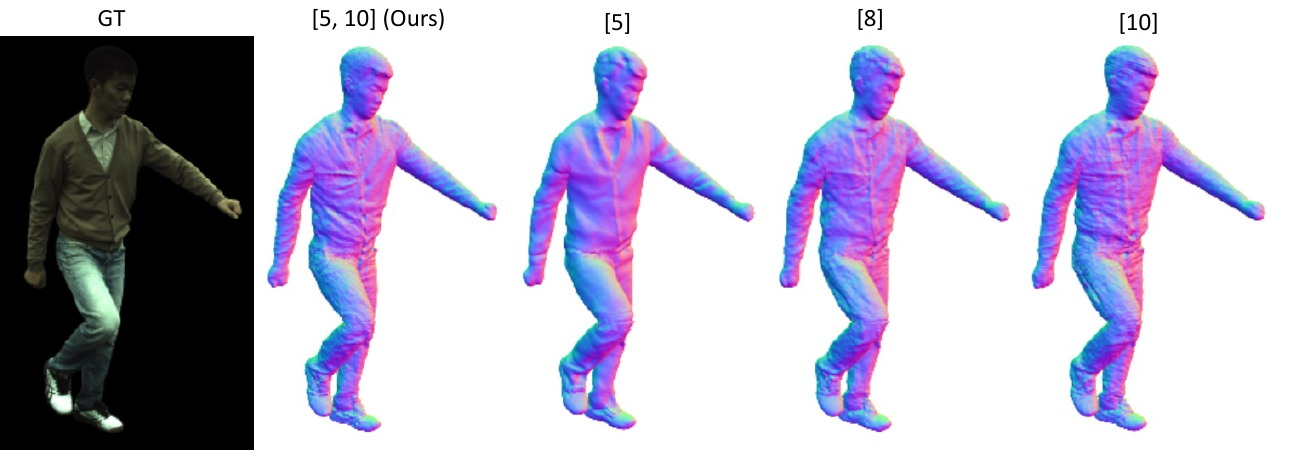}
    \caption{
    \cj{
    \textbf{Pose-dependent branch with different frequency bands.} 
    The model with only pose-dependent branch is prone to either blur fine-grained details (e.g. [5]), or introduce unwanted noise (e.g. [10]).
    In contrast, our model with dual branches can successfully reproduce multi-scale signals.
    }
    }
    \label{fig:supp-ab-hf}
\end{figure*}

\begin{figure*}[hb]
    \centering
    \includegraphics[width=0.9\linewidth,trim={0 20cm 0cm 0cm},clip]{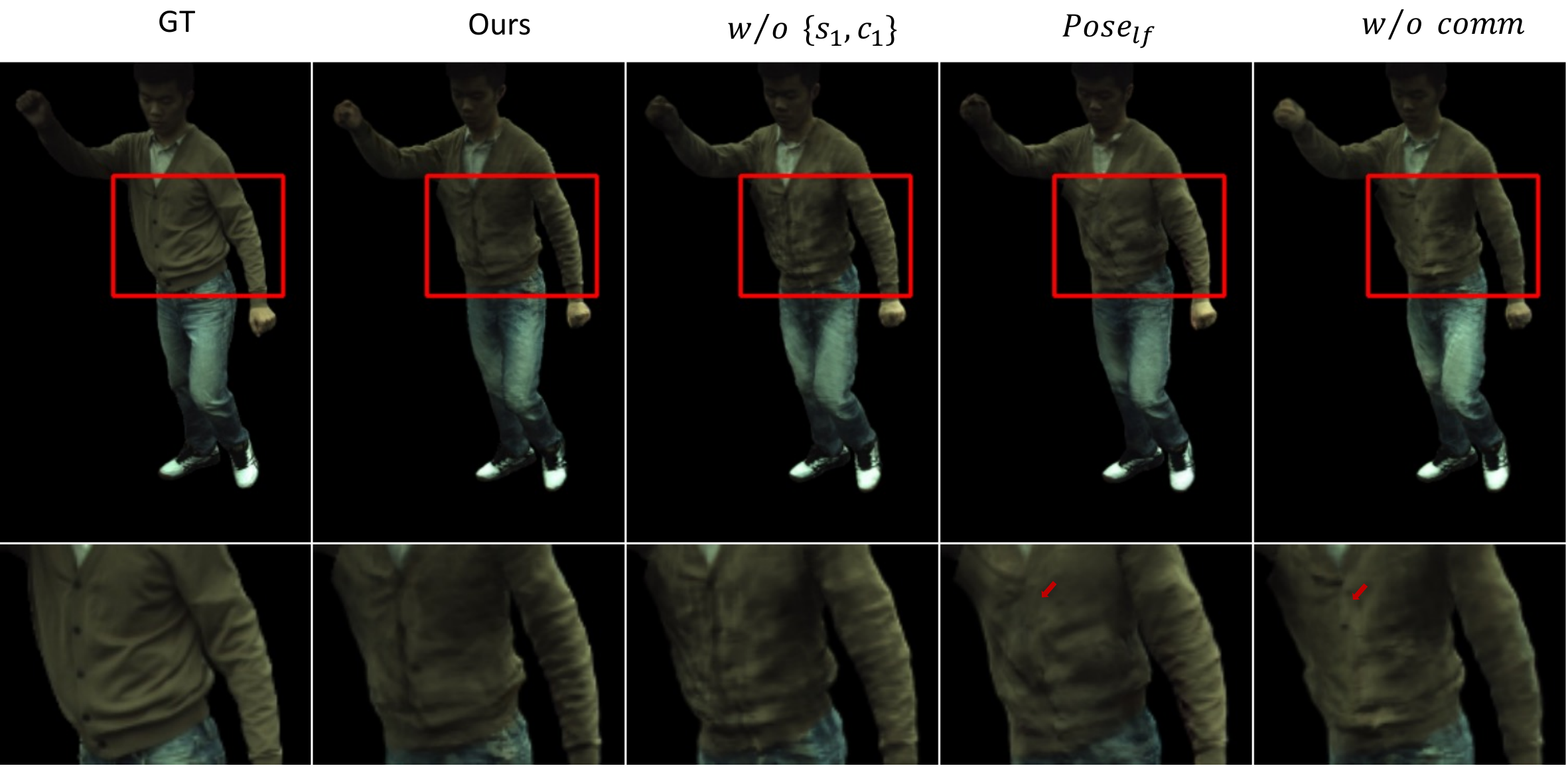}
    \caption{
    \textbf{Ablation study on pose-independent deformation outputs.}
    Together with the ablated baseline ($\mathrm{w/o\ comm}$) which only outputs $\{s_2, c_2\}$, other two ablation models ($\mathrm{w/o\ \{s_1, c_1\}}$ and $\mathrm{Pose}_{lf}$) all disenable explicit modeling of pose-independent deformation outputs.
    Thus they either blurs the significant textures (e.g. highlighted stripe pattern by red arrows), or introduce heavily unwanted black artifacts (e.g. $\mathrm{w/o\ \{s_1, c_1\}}$).
    }
    \label{fig:ab_hf_poself_comm}
\end{figure*}

\textbf{3.}~Explicit modeling of pose-independent deformation outputs plays an important role in our framework.
We can see in Fig.~\ref{fig:ab_hf_poself_comm} that, both $\mathrm{w/o\ \{s_1, c_1\}}$ and $\mathrm{Pose}_{lf}$ only depend on pose-dependent outputs and thus blur or distort the desired textures.
In Fig.~\ref{fig:supp-ab-hf}, Tab.~\ref{tab:mocap-consistency} and Tab.~\ref{tab:supp_ablation_freq}, we present the results of training only pose-dependent branch with different frequencies.
Here we denote the model with $L_d=X$ as $[X]$.
We can see that, simply adjusting $L_d$ cannot successfully capture multi-scale patterns.
Moreover, to verify the effectiveness of disentangling in output space, we train a new baseline by keeping the full framework but only applying $\{s_2, c_2\}$ for rendering and denote this baseline as $\mathrm{w/o\ comm}$.
In the last column of Fig.~\ref{fig:ab_hf_poself_comm}, the $\mathrm{w/o\ comm}$ baseline blurs the textures without modeling pose-independent outputs.

\begin{figure*}[t]
    \centering
    \includegraphics[width=1.\linewidth,trim={0 10cm 0cm 0cm},clip]{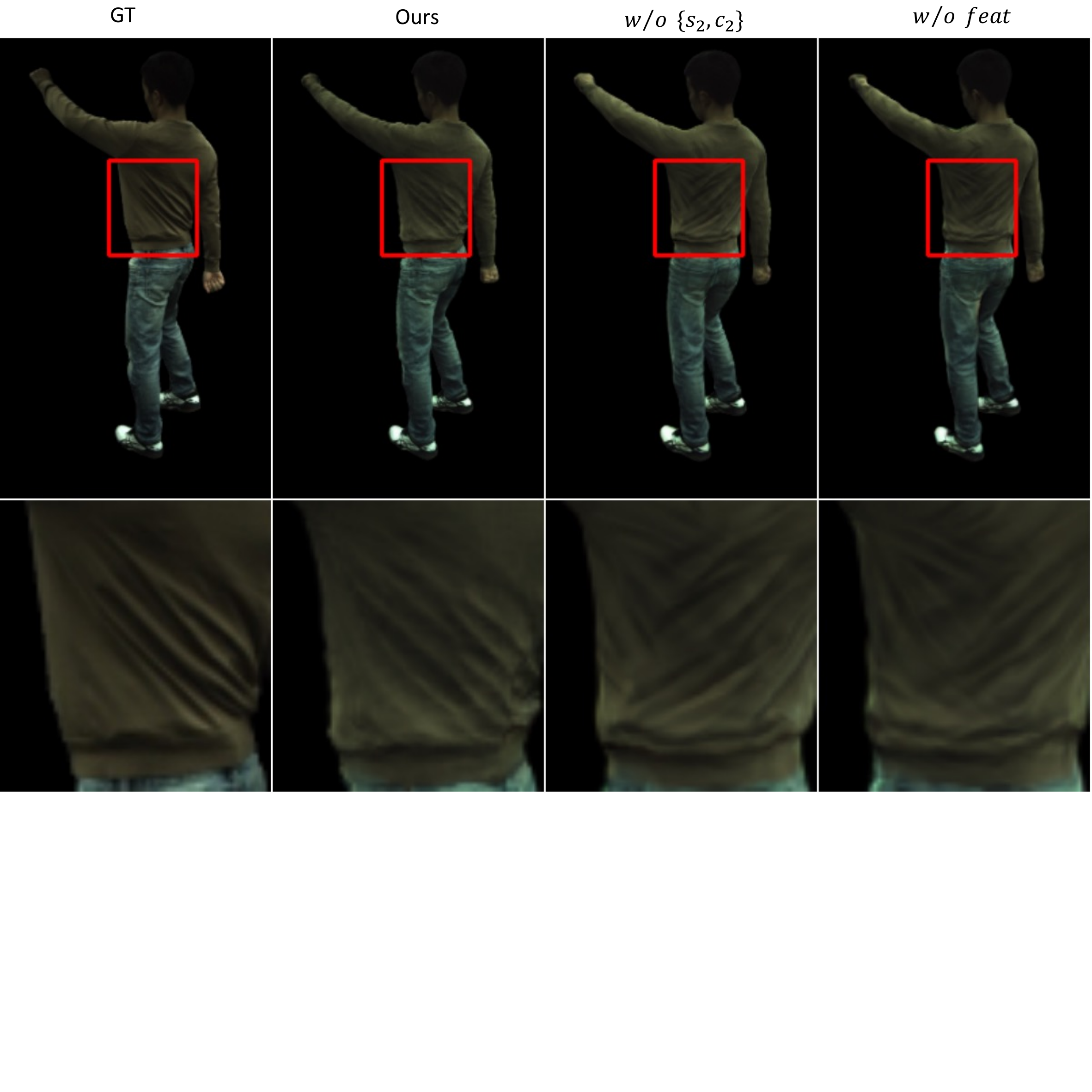}
    \caption{
    \textbf{Ablation study on the pose-dependent deformation outputs.}
    Although the two ablated models, which remove the pose-dependent deformation outputs, can also generate intricate wrinkles, our complete model produces superior textures with more accurate pattern directions.
    }
    \label{fig:ab_lf_feat}
\end{figure*}

\textbf{4.}~Correspondingly, we present the ablation results on the pose-dependent deformation outputs in Fig.~\ref{fig:ab_lf_feat}. 
While the ablated baselines successfully synthesize cloth wrinkles, our full model can generate more adaptive texture patterns.

\begin{figure*}[p]
    \centering
    \includegraphics[width=0.85\linewidth,trim={0 9cm 21.4cm 0cm},clip]{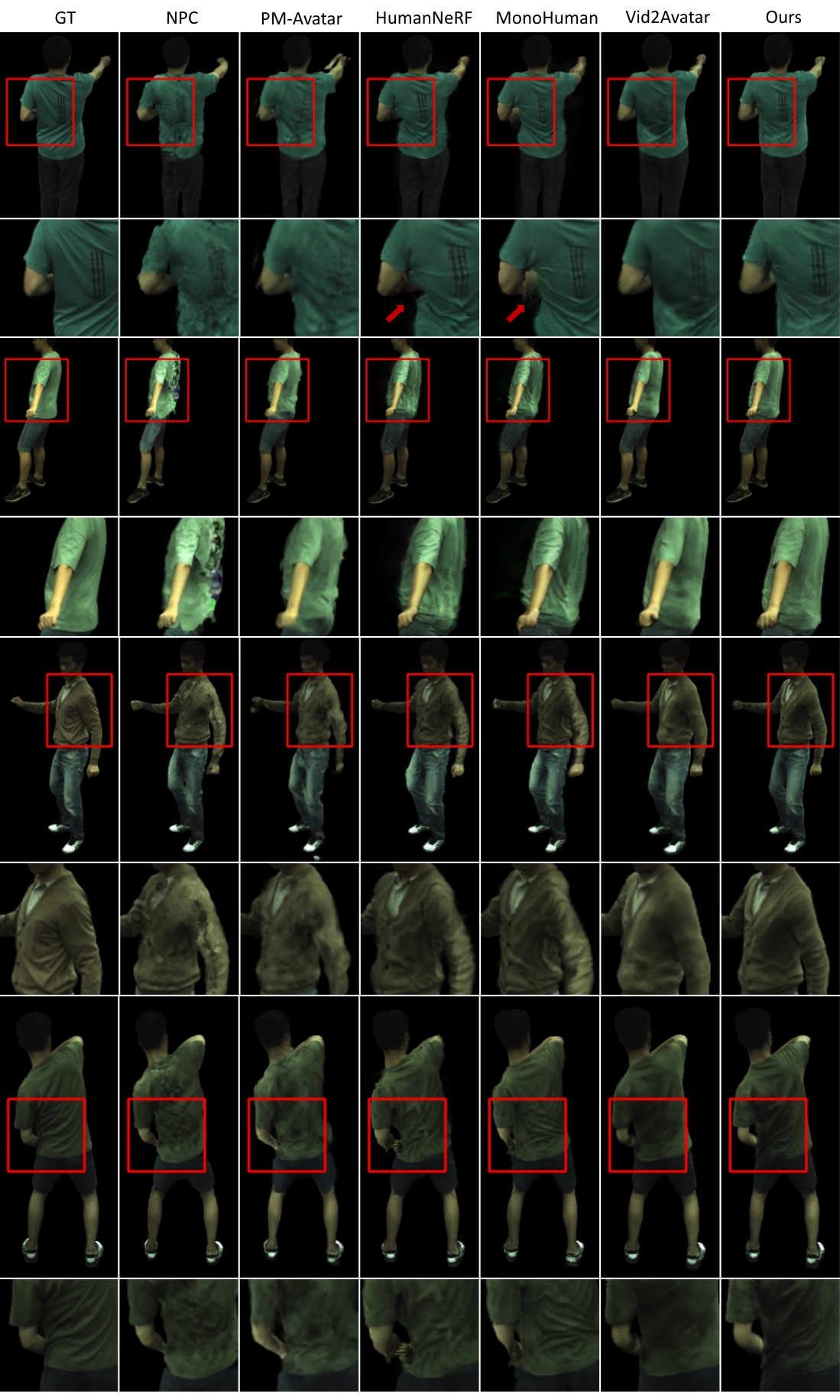}
    \caption{
    \cj{
    \textbf{Novel view synthesis on ZJU-Mocap.} 
    We can preserve superior sharp contours and synthesize more adaptive textures than baselines.
    }
    }
    \label{fig:supp-mocap-nvs}
\end{figure*}
\begin{figure*}[p]
    \centering
\includegraphics[width=0.85\linewidth,trim={0 9cm 21.4cm 0cm},clip]{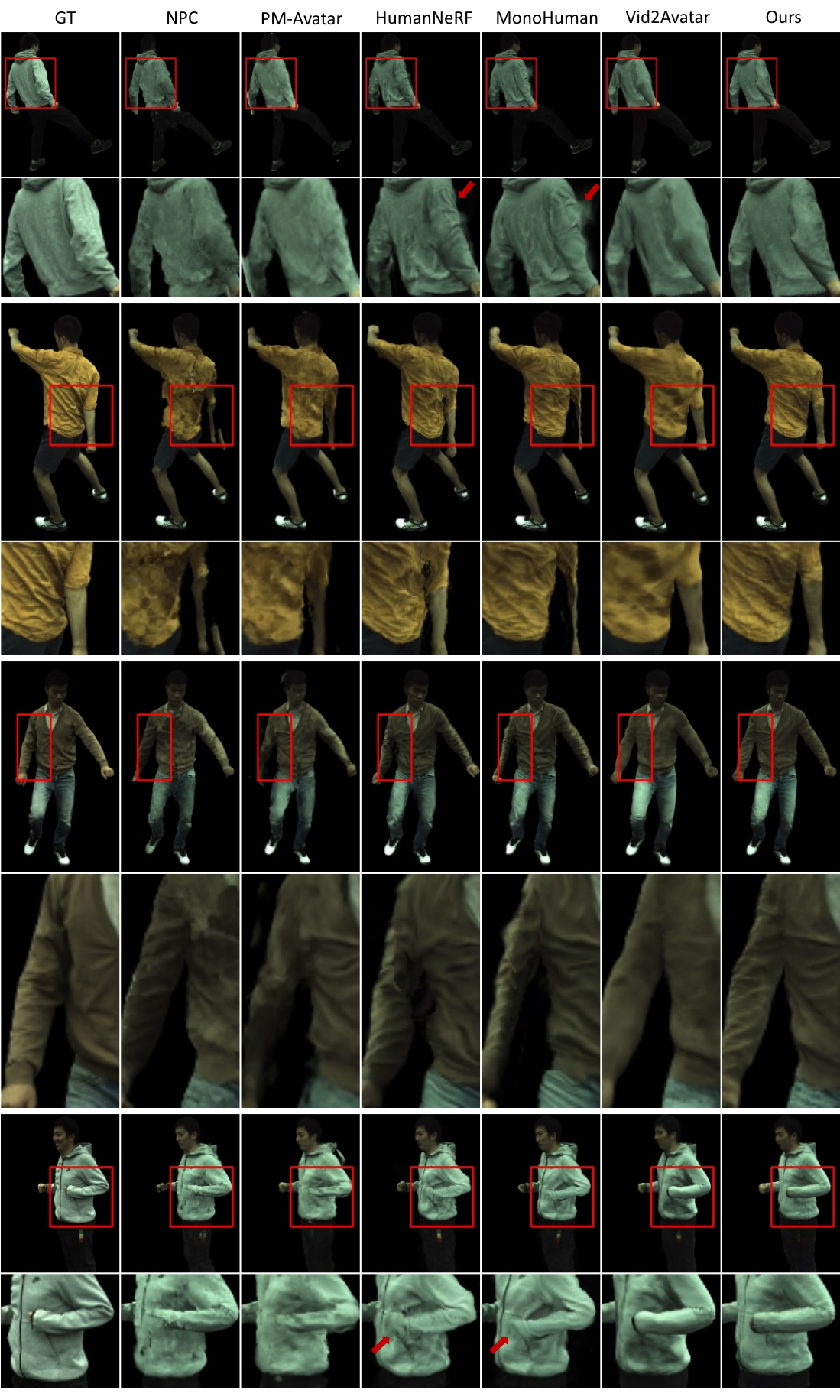}
    \vspace{-1.em}
    \caption{
    \cj{
    \textbf{Novel pose rendering on ZJU-Mocap.} 
    Compared to baselines, we can better generate stable body boundaries without blurry artifacts.
    }
    }
    \label{fig:supp-mocap-npr}
\end{figure*}

\section{More Results}
\label{sec:supp_result}

\cj{
\begin{figure*}[t!]
    \centering
    \includegraphics[width=0.9\linewidth,trim={0 6cm 0cm 0cm},clip]{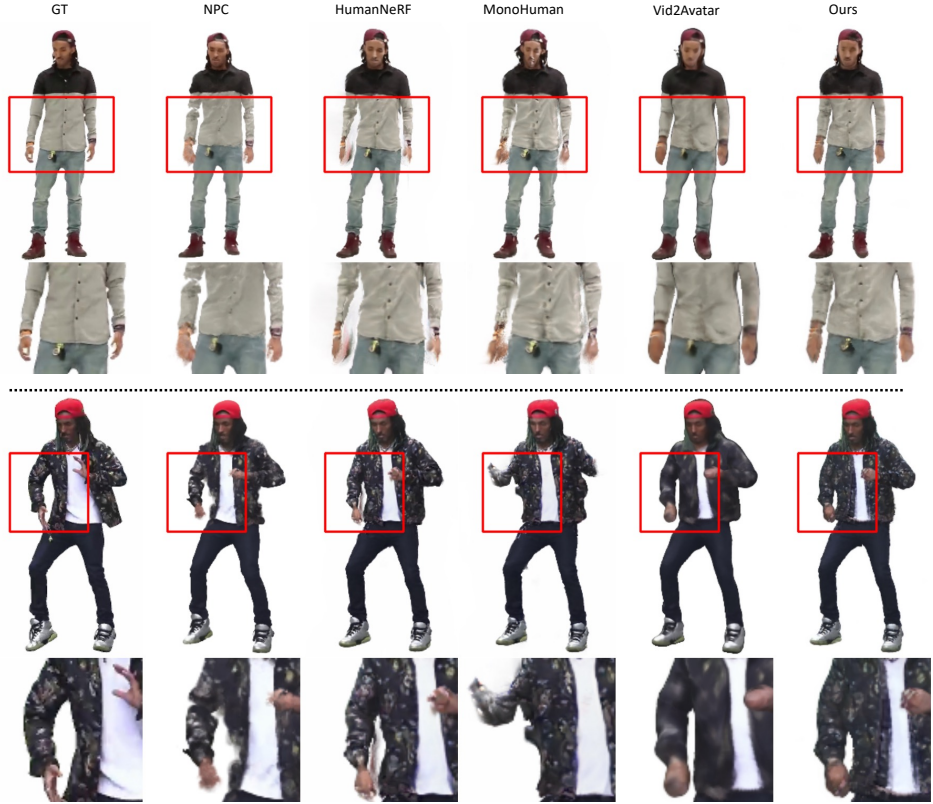}
    \caption{
    \textbf{Novel pose rendering on Youtube sequences.}
    \cj{Our method can preserve better shape contours and produce more realistic cloth textures(e.g. the buttons shown in $1^st$ row).}
    }
    \label{fig:youtube-npr-full}
\end{figure*}

We provide visual comparisons for more ZJU-Mocap sequences to highlight our generality.
In Fig.~\ref{fig:supp-mocap-nvs} and Fig.~\ref{fig:supp-mocap-npr}, we visualize the rendering results of novel view synthesis and novel pose rendering respectively.
Compared to baselines, our method can better preserve sharp body contours and adaptive texture details in both tasks.
}

\begin{figure*}[ht]
    \centering
    \includegraphics[width=0.95\linewidth,trim={0 28cm 0cm 0cm},clip]{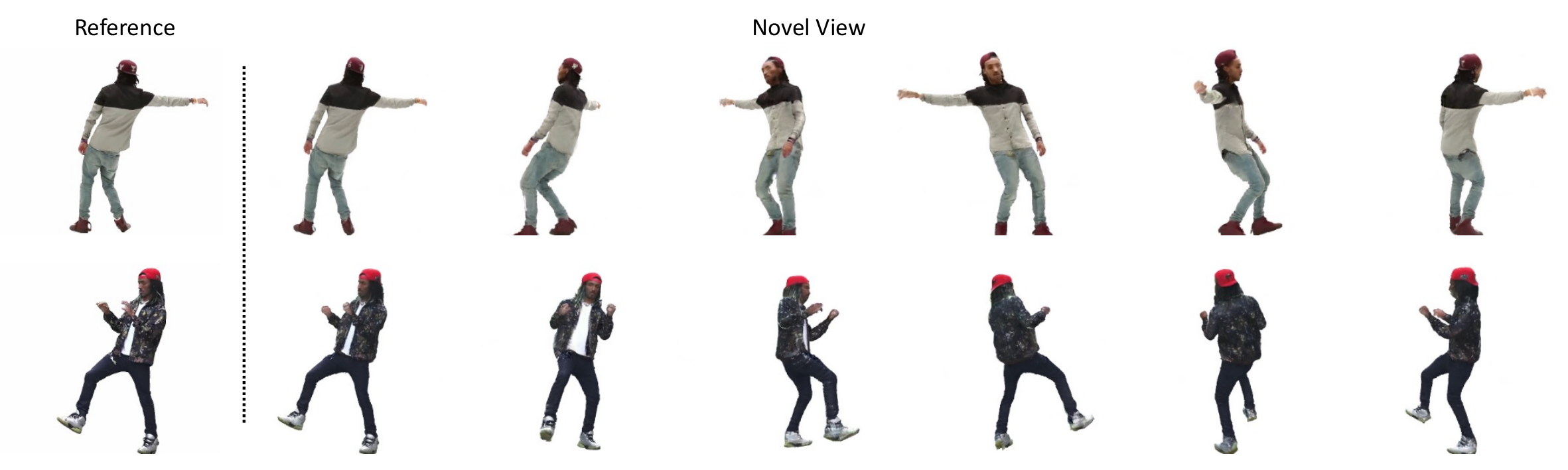}
    \caption{
    \textbf{Novel View Synthesis on Youtube sequences.}
    Our method can successfully generalize to novel camera views.
    }
    \label{fig:youtube-nvs}
\end{figure*}

\cj{
In Fig.~\ref{fig:youtube-npr-full}, we illustrate one more example for the novel pose rendering on Youtube test sets. 
Similar to Fig.~\ref{fig:youtube-npr} in the main text, our method can produce textured patterns on coat without obvious distortions.
}
Following HumanNeRF~\cite{weng2022humannerf} and MonoHuman~\cite{yu2023monohuman}, we produce the novel view synthesis results for Youtube sequences.
Fig.~\ref{fig:youtube-nvs} shows that, our method is capable of producing high fidelity details similar to the ground truth even on completely unobserved views.

\begin{figure*}[t]
    \centering
    \includegraphics[width=0.9\linewidth,trim={0 27cm 0cm 0cm},clip]{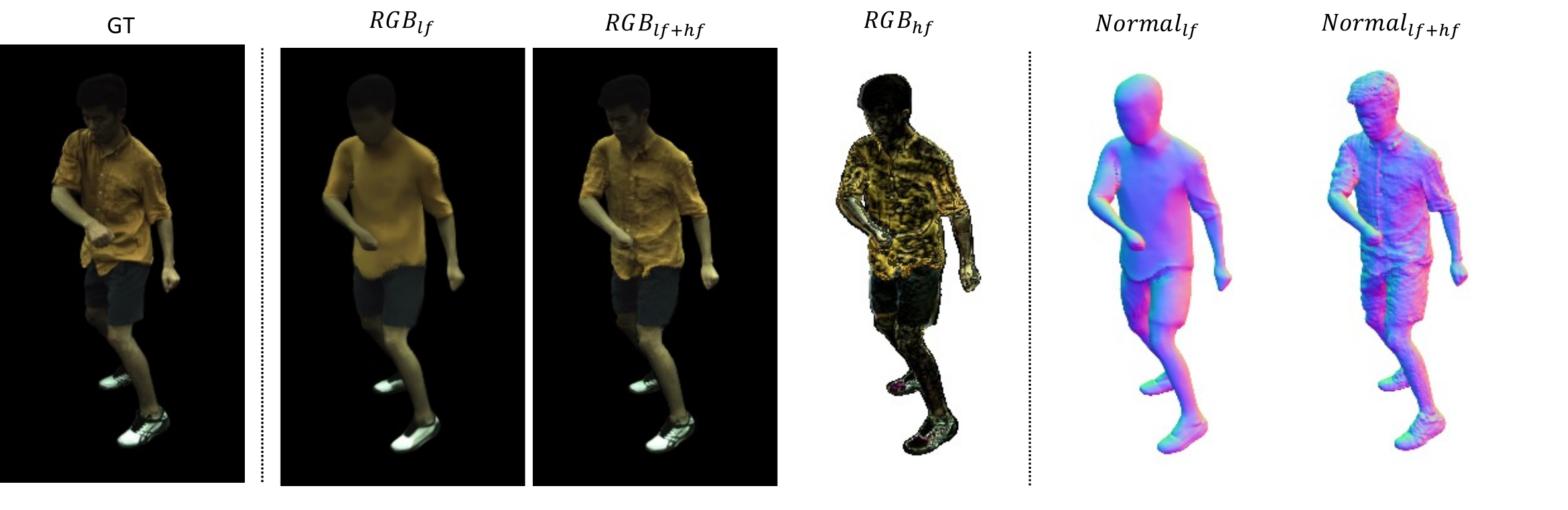}
    \caption{
    \cj{
    \textbf{Outputs from two branches.} 
    Adhering to our network design, the pose-independent branch outputs the low-frequency base appearance as $\mathrm{RGB}_{lf}$ while the pose-dependent branch estimates the corresponding high-frequency residual as $\mathrm{RGB}_{hf}$. Combining $\mathrm{RGB}_{lf}$ with $\mathrm{RGB}_{hf}$ reconstructs image with all frequencies as $\mathrm{RGB}_{lf+hf}$. Similarly, the normal map from pose-independent branch ($\mathrm{Normal}_{lf}$) presents overall smooth patterns while our full normal output ($\mathrm{Normal}_{lf+hf}$) has realistic geometric patterns. $\mathrm{RGB}_{hf}$ is scaled for better visualizations. 
    }
    \vspace{-0.5em}
    }
    \label{fig:supp_mocap_branch}
\end{figure*}

\cj{
Fig.~\ref{fig:supp_mocap_branch} visualizes the normal maps from two branches.
Although the pose-independent branch succeeds in generating a reasonable body boundary, 
the output exhibits over-smooth patterns without high-fre\-quency details.
In contrast, our full model with both branches preserves the perceptive outlines and adaptive pose-dependent textures simultaneously, which supports our conceptual motivation.
}

\begin{figure*}[t!]
    \centering
    \includegraphics[width=0.9\linewidth,trim={0 18cm 18cm 0cm},clip]{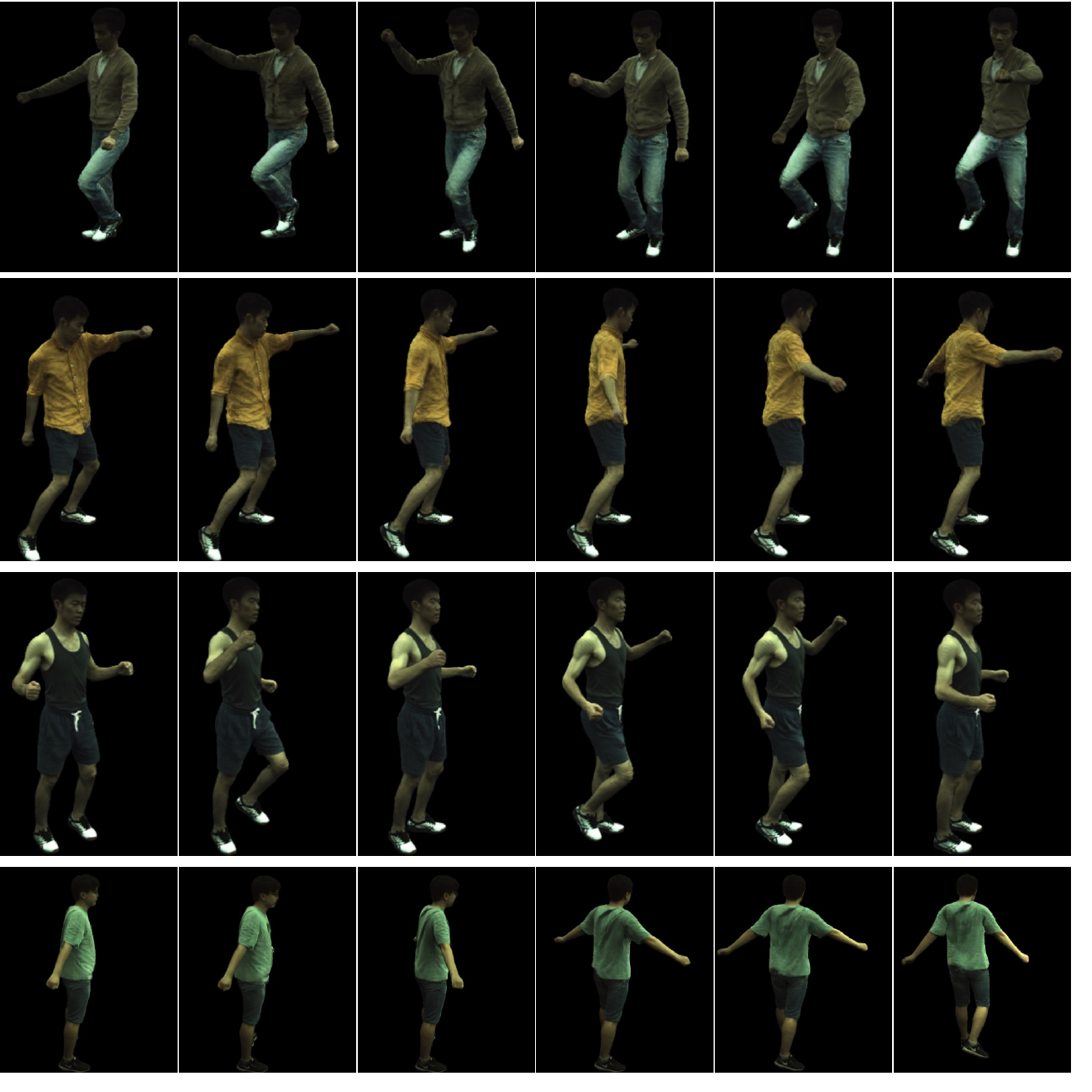}
    \caption{
    \textbf{Image Gallery on ZJU-MoCap dataset.}
    Our method maintains strong time consistency across various avatars.
    }
    \label{fig:image_gallery}
\end{figure*}

Preserving time consistency is a critical aspect for 3D avatar rendering and human animations. 
To achieve this, we also generate a collection of images for each avatar featuring various human poses.
In Fig.~\ref{fig:image_gallery}, we can see that our network is robust to various poses across different avatars and accordingly produces stable results, showing reliable time consistency.
See the attached video for more clear demonstration.

\cj{
Finally, we apply the optical flow~\cite{horn1981determining} to measure the frame consistency scores and report the numbers in Tab.~\ref{tab:mocap-consistency}.
Compared to previous baselines which neglect modeling pose-independent outputs explicitly, our method benefits from significant higher video consistency, which can partially explain our generalization to novel poses.
As a key network design, our frequency assumption on the pose-independent component also helps stablize the output frames.
Moreover, complying with the discussions in Sec.~\ref{sec:supp_abtest}, training with one branch is inferior in getting better frame consistency.
}

\begin{figure*}[t!]
    \centering
    \includegraphics[width=0.9\linewidth,trim={0 28cm 7cm 0cm},clip]{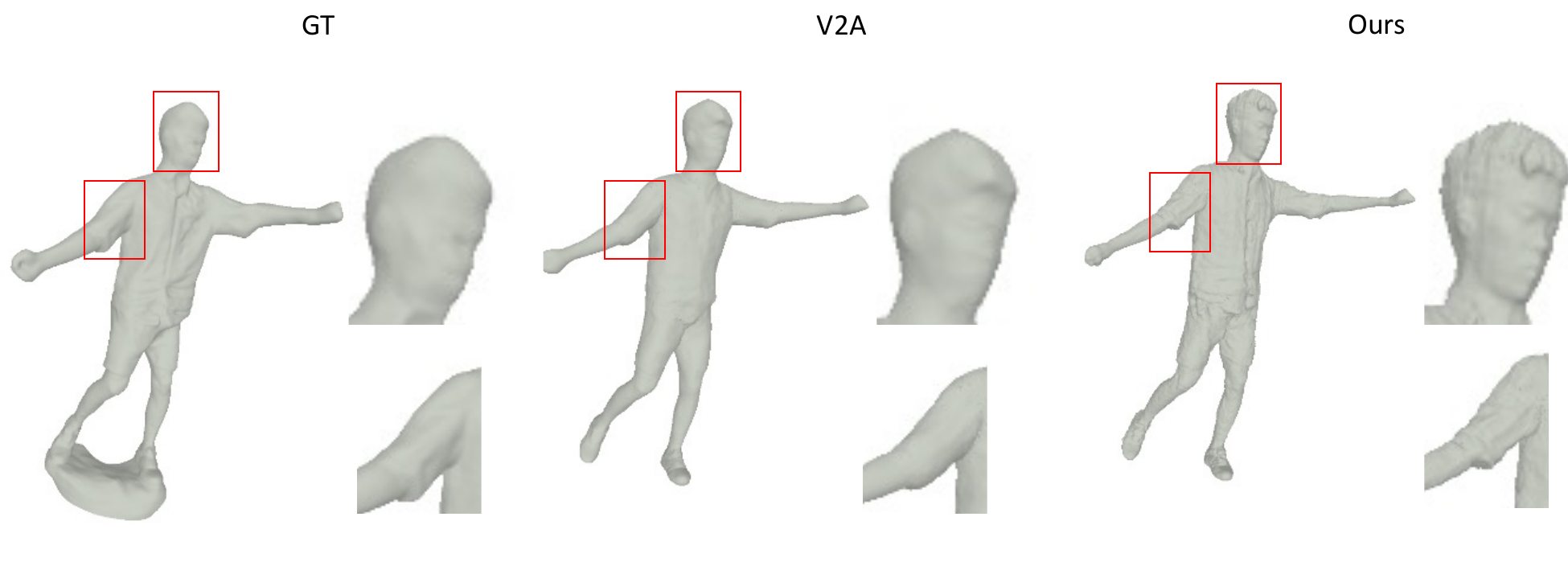}
    \caption{
    \textbf{Visualization for pseudo ground truth shape.}
    Both shapes from pseudo ground truth (GT) and Vid2Avatar (V2A) smooth over the geometric details (e.g. the highlighted facial expressions and shoulder). 
    Moreover, the pseudo GT shape also introduces the unwanted ground surfaces on the bottom, benefiting Vid2Avatar more from computing quantitative metrics.
    }
    \label{fig:mocap_shape_gt}
\end{figure*}

\cj{
\section{Complete Metrics on ZJU-Mocap sequences}
\label{sec:supp_mocap_metrics}

Besides the overall average numbers in the Tab.~\ref{tab:mocap-score} of the main text, we also report a per-subject breakdown of the quantitative metrics against all baseline methods. Specifically, Tab.~\ref{tab:mocap-novel-view} lists the scores for the novel view synthesis while Tab.~\ref{tab:mocap-novel-pose} details each
method’s results in novel pose rendering. Being consistent with the visual results shown in Fig.~\ref{fig:mocap} of the main text, our method almost outperforms all baselines for all subjects.
}

\begin{table}[b]
\caption{
\textbf{Novel-view synthesis comparisons on ZJU-Mocap~\cite{peng2021neural}}. 
Our method achieves better novel view synthesis with about 10\% improvement in LPIPS and KID than the best baseline.
}
\centering
\resizebox{1.0\linewidth}{!}{
\setlength{\tabcolsep}{3pt}
\begin{tabular}{lcccccccccccccccccccc}
\toprule
 & \multicolumn{4}{c}{HumanNeRF~\cite{weng2022humannerf}}  & \multicolumn{4}{c}{MonoHuman~\cite{yu2023monohuman}} & \multicolumn{4}{c}{NPC~\cite{su2023npc}} & \multicolumn{4}{c}{Vid2Avatar~\cite{guo2023vid2avatar}} & \multicolumn{4}{c}{\textbf{Ours}} \\
 \midrule
 & PSNR~$\uparrow$& SSIM~$\uparrow$& LPIPS~$\downarrow$& KID~$\downarrow$& PSNR~$\uparrow$& SSIM~$\uparrow$& LPIPS~$\downarrow$& KID~$\downarrow$& PSNR~$\uparrow$& SSIM~$\uparrow$& LPIPS~$\downarrow$& KID~$\downarrow$& PSNR~$\uparrow$& SSIM~$\uparrow$& LPIPS~$\downarrow$& KID~$\downarrow$& PSNR~$\uparrow$& SSIM~$\uparrow$& LPIPS~$\downarrow$& KID~$\downarrow$\\
\rowcolor{gray}
S313 & 29.49 & 0.965 & 30.46 & 9.43 & 29.55 & \textbf{0.966} & 31.40 & \textbf{7.25} & \textbf{29.70} & \textbf{0.966} & 36.29 & 54.37 & 28.44 & \textbf{0.966} & 37.74 & 30.08 & 28.79 & \textbf{0.966}& \textbf{29.61} & 10.42\\
S377 & 30.15 & 0.974 & 23.17 & 5.37 & 30.22 & 0.976 & 23.70 & 3.95 & \textbf{30.52} & \textbf{0.977} & 22.29 & 4.75 & 29.52 & 0.975 & 24.91 & 12.52 & 29.92 & \textbf{0.977}& \textbf{19.85} & \textbf{3.47} \\
\rowcolor{gray}
S386 & 32.69 & 0.973 & 31.99 & 31.06 & 32.77 & 0.970 & 36.69 & 28.08 & 32.07 & 0.973 & 36.88 & 79.70 & \textbf{33.04} & \textbf{0.977} & 30.65 & 23.42 & 33.00 & \textbf{0.977}& \textbf{27.62} & \textbf{15.45}\\
S387 & 27.95 & 0.962 & 34.85 & 15.70 & \textbf{28.20} & 0.962 & 37.99 & 21.99 & 27.76 & 0.962 & 43.34 & 80.01 & 28.15 & \textbf{0.965} & 40.26 & 25.17 & 28.09 & 0.964 & \textbf{33.70} & \textbf{14.72}\\
\rowcolor{gray}
S390 & 30.16 & 0.968 & 33.09 & 19.53 & 30.27 & 0.968 & 34.68 & 20.00 & 29.74 & 0.963 & 46.02 & 71.72 & \textbf{30.73} & \textbf{0.970} & 37.33 & 22.27 & 30.66 & \textbf{0.970} & \textbf{32.86} & \textbf{15.51}\\
S392 & 30.70 & 0.970 & 32.02 & 9.88 & 30.94 & 0.971 & 32.30 & \textbf{6.26} & \textbf{31.70} & 0.972 & 33.34 & 38.44 & 30.46 & 0.971 & 36.58 & 31.50 & 31.18 & \textbf{0.973}& \textbf{29.17} & 10.31\\
\rowcolor{gray}
S393 & 28.38 & 0.961 & 34.95 & 12.67 & 28.40 & 0.961 & 36.64 & \textbf{10.20} & \textbf{28.61} & \textbf{0.963} & 40.88 & 55.83 & 27.94 & 0.962 & 40.93 & 39.11 & 28.49 & \textbf{0.963}& \textbf{34.62} & 14.83\\
S394 & 30.00 & 0.962 & 33.98 & 10.18 & 30.02 & 0.963 & 34.38 & \textbf{7.64} & 30.07 & 0.963 & 38.38 & 41.14 & 29.82 & 0.965 & 36.46 & 37.11 & \textbf{30.68} & \textbf{0.967}& \textbf{29.74} & 9.07\\
\midrule
Avg & 29.94 & 0.967 & 31.81 & 14.23 & 30.03 & 0.967 & 33.47 & 13.18 & 30.01 & 0.967 & 37.18 & 53.24 & 29.76 & 0.969 & 35.61 & 27.65 & \textbf{30.11}& \textbf{0.970}& \textbf{29.64} & \textbf{11.72}\\
\bottomrule
\end{tabular}%
\label{tab:mocap-novel-view}%
}%
\end{table}%

\begin{table}[t]
\caption{
\textbf{Novel-pose synthesis comparisons on ZJU-Mocap~\cite{peng2021neural}}. 
Our method enables better generalization to unseen poses with about 12\% improvement in LPIPS and KID than the best baseline, experimentally supporting our frequency-aware field factorization design. 
}
\centering
\resizebox{1.0\linewidth}{!}{
\setlength{\tabcolsep}{3pt}
\begin{tabular}{lcccccccccccccccccccc}
\toprule
 & \multicolumn{4}{c}{HumanNeRF~\cite{weng2022humannerf}}  & \multicolumn{4}{c}{MonoHuman~\cite{yu2023monohuman}} & \multicolumn{4}{c}{NPC~\cite{su2023npc}} & \multicolumn{4}{c}{Vid2Avatar~\cite{guo2023vid2avatar}} & \multicolumn{4}{c}{\textbf{Ours}} \\
 \midrule
 & PSNR~$\uparrow$& SSIM~$\uparrow$ & LPIPS~$\downarrow$& KID~$\downarrow$& PSNR~$\uparrow$& SSIM~$\uparrow$& LPIPS~$\downarrow$& KID~$\downarrow$& PSNR~$\uparrow$& SSIM~$\uparrow$& LPIPS~$\downarrow$& KID~$\downarrow$& PSNR~$\uparrow$& SSIM~$\uparrow$& LPIPS~$\downarrow$
 & KID~$\downarrow$& PSNR~$\uparrow$& SSIM~$\uparrow$& LPIPS~$\downarrow$ & KID~$\downarrow$\\
\rowcolor{gray}
S313 & 28.08 & 0.961 & 32.37 & 12.17 & 28.24 & 0.963 & 32.81 & \textbf{8.02} & 27.41 & 0.963 & 37.13 & 53.82 & 28.00 & 0.965 & 38.43 & 32.55 & \textbf{28.41}& \textbf{0.967}& \textbf{29.01} & 10.85\\
S377 & 30.12 & 0.976 & 22.68 & 5.63 & 30.23 & 0.977 & 23.87 & 3.31 & \textbf{30.61} & \textbf{0.978} & 21.55 & 4.42 & 29.80 & 0.977 & 24.77 & 15.94 & 30.22 & \textbf{0.978}& \textbf{19.25} & \textbf{3.26}\\
\rowcolor{gray}
S386 & 32.11 & 0.970 & 34.48 & 25.23 & 32.41 & 0.968 & 39.45 & 25.10 & 31.71 & 0.971 & 37.95 & 68.22 & 32.48 & \textbf{0.975} & 32.43 & 24.64 & \textbf{32.91}& \textbf{0.975}& \textbf{28.51} & \textbf{10.80}\\
S387 & 27.55 & 0.962 & 34.67 & 17.47 & 27.61 & 0.961 & 38.34 & 23.90 & 27.24 & 0.962 & 42.14 & 83.10 & 27.34 & \textbf{0.964} & 40.83 & 29.66 & \textbf{27.75}& \textbf{0.964}& \textbf{32.42} & \textbf{17.38}\\
\rowcolor{gray}
S390 & 30.01 & 0.967 & 32.24 & \textbf{12.43} & 30.62 & 0.968 & 35.84 & 17.55 & 30.02 & 0.965 & 42.60 & 60.21 & 30.81 & 0.971 & 34.73 & 28.09 & \textbf{31.10}& \textbf{0.972}& \textbf{28.91} & 16.79\\
S392 & 30.61 & 0.970 & 32.08 & \textbf{5.93} & 30.88 & 0.971 & 33.52 & 5.09 & \textbf{31.33} & \textbf{0.972} & 34.88 & 44.54 & 30.50 & 0.971 & 37.53 & 46.91 & 30.95 & \textbf{0.972}& \textbf{29.80} & 10.79\\
\rowcolor{gray}
S393 & 28.36 & 0.961 & 34.31 & \textbf{10.70} & 28.75 & \textbf{0.963} & 35.43 & 9.46 & 28.76 & \textbf{0.963} & 39.03 & 51.61 & 28.09 & 0.962 & 40.17 & 38.36 & \textbf{28.81}& \textbf{0.963}& \textbf{31.71} & 11.26\\
S394 & 28.78 & 0.960 & 34.64 & 8.95 & 29.10 & 0.961 & 34.75 & 8.41 & \textbf{29.77} & 0.963 & 36.90 & 32.45 & 29.24 & 0.963 & 36.61 & 35.94 & 29.69 & \textbf{0.965}& \textbf{29.41} & \textbf{7.61}\\
\midrule
Avg & 29.45 & 0.966 & 32.18 & 12.32 & 29.73 & 0.967 & 34.25 & 12.61 & 29.61 & 0.967 & 36.52 & 49.79 & 29.53 & 0.969 & 35.69 & 31.51 & \textbf{29.98}& \textbf{0.970}& \textbf{28.60} & \textbf{11.09}\\
\bottomrule
\end{tabular}
\label{tab:mocap-novel-pose}
}
\end{table}

\section{Comparisons on 3D Shape Reconstructions}
\label{sec:supp_3d_recon}

\begin{figure*}[t!]
    \centering
    \includegraphics[width=0.8\linewidth,trim={0 30cm 1.2cm 0cm},clip]{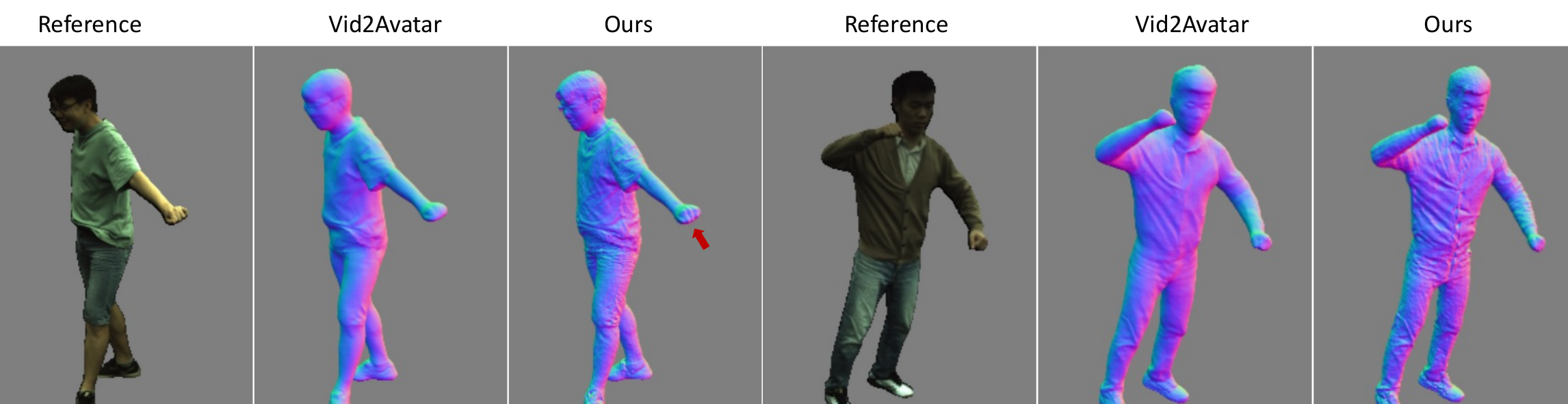}
    \caption{
    \textbf{Qualitative Comparisons for Normal Map.}
    Compared to Vid2Avatar, our method can generate more significant textures with faithful details, such as the fist pattern in the left example \& the avatar face in the right example.
    }
    \label{fig:mocap-normal-map}
\end{figure*}

\cj{
\begin{figure*}[ht]
    \centering
    \includegraphics[width=0.9\linewidth,trim={0 14cm 0cm 0cm},clip]{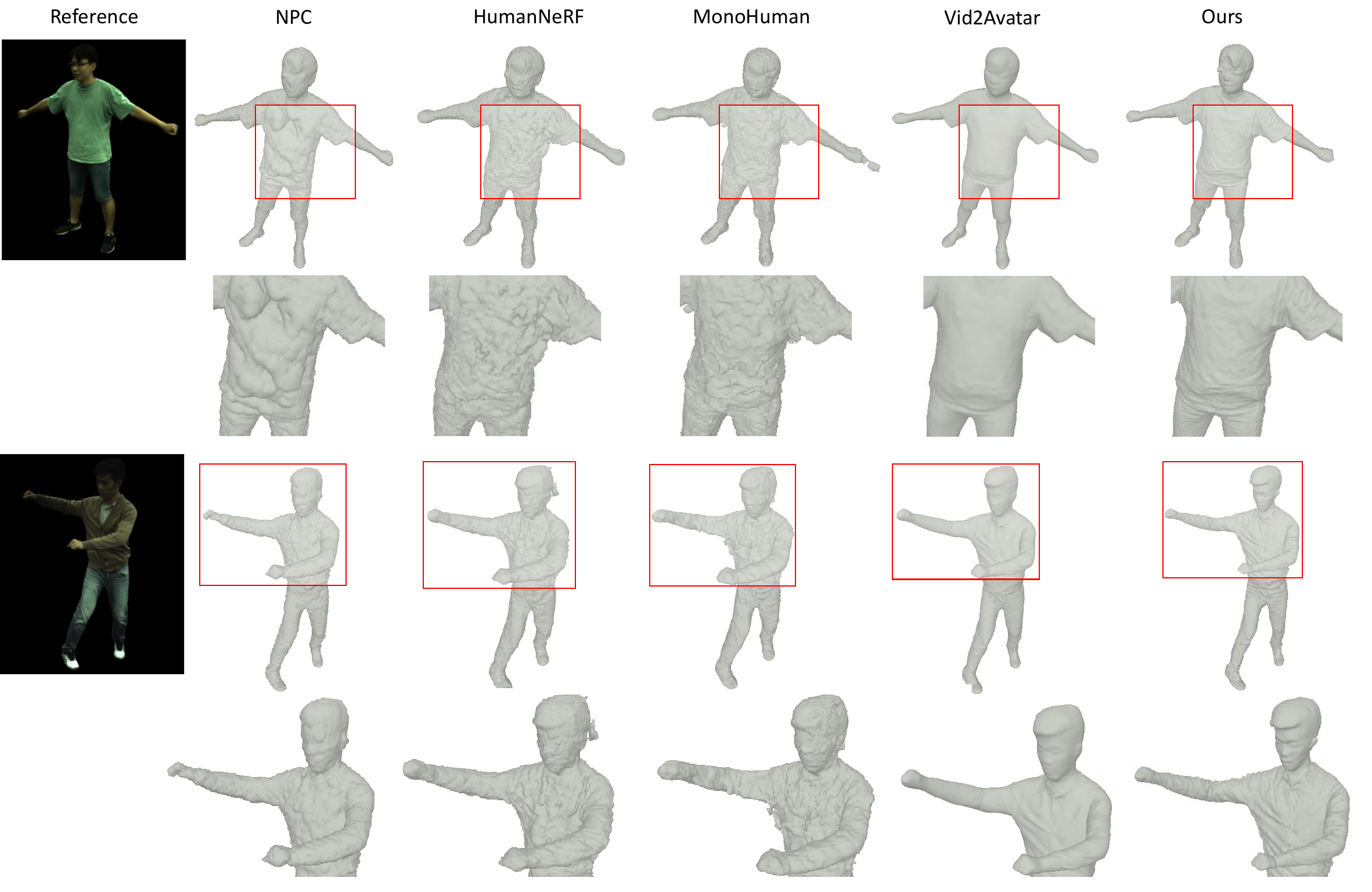}
    \caption{
    \textbf{Comparisons on geometry reconstruction.} 
    Our method yields more precise shape estimates with fine-grained geometric details.
    }
    \label{fig:mocap-geometry}
\end{figure*}

Besides Fig.~\ref{fig:mocap-geometry} in the main text, we present visualization results for the extracted 3D meshes on another ZJU-Mocap sequence in Fig.~\ref{fig:mocap-geometry}.
Compared to baselines, our method yields more structured geometric details in the highlighted area while baselines either over-smooth the wrinkles or introduce unwanted bump noise on coat.

In Fig.~\ref{fig:mocap-normal-map}, we provide the produced normal maps to additionally show our empirical advantages in producing detailed geometric textures over Vid2Avatar.
Complying with the rendering results of RGB images, our frequency-aware field factorization creates geometric patterns in various scales.
}

Then we further report the quantitative evaluations for each used sequence in Tab.~\ref{tab:mocap-geomtry}.
Respecting the $L_2$ Chamfer Distance (CD) and Normal Consistency (NC), our method can achieve the on-par or even best result across almost all six sequences.
In detail, compared to the density-based baselines, including HumanNeRF, MonoHuman and NPC, we yield much lower \textbf{CD} and much higher \textbf{NC} scores.
For Vid2Avatar which also uses SDF for volume rendering, we still possess superior \textbf{CD} results due to our more faithful geometric textures.
However, as shown in Fig.~\ref{fig:mocap_shape_gt}, the pseudo ground truth shape reconstructed as ARAH~\cite{wang2022arah} smooths out the surface details and introduce unwanted artifacts (e.g. the ground surfaces).
Thus the imperfect pseudo ground truth shapes benefit Vid2Avatar in most sequences with a slightly higher \textbf{Normal Consistency} metrics than ours.
How to more accurately evaluate the quality of reconstructed 3D shapes for NeRF-based methods is still an open problem and will be our future research direction.

\begin{table}[!t]
\caption{
\textbf{Geometry reconstruction comparisons on ZJU-Mocap~\cite{peng2021neural}}. 
We report L2 Chamfer Distance (CD) and Normal Consistency (NC) to demonstrate our capability in synthesizing high-quality shapes with fine-grained details. 
Note that the imperfect pseudo-ground-truth meshes prevent us from achieving more significant quantitative improvements. 
}
\centering
\resizebox{1.0\linewidth}{!}{
\setlength{\tabcolsep}{3pt}
\begin{tabular}{lcccccccccc}
\toprule
 & \multicolumn{2}{c}{HumanNeRF~\cite{weng2022humannerf}}  & \multicolumn{2}{c}{MonoHuman~\cite{yu2023monohuman}} & \multicolumn{2}{c}{NPC~\cite{su2023npc}} & \multicolumn{2}{c}{Vid2Avatar~\cite{guo2023vid2avatar}} & \multicolumn{2}{c}{\textbf{Ours}} \\
 \midrule
 & CD~$\downarrow$& NC~$\uparrow$ & CD~$\downarrow$& NC~$\uparrow$ & CD~$\downarrow$& NC~$\uparrow$ & CD~$\downarrow$& NC~$\uparrow$ & CD~$\downarrow$& NC~$\uparrow$ \\
\rowcolor{gray}
S313 & 0.197 & 0.661 & 0.289 & 0.637 & 0.039 & 0.824 & 0.017 & \textbf{0.916} & \textbf{0.014} & 0.904 \\
S377 & 0.307 & 0.630 & 0.368 & 0.616 & 0.046 & 0.838 & 0.036 & 0.859 & \textbf{0.020} & \textbf{0.894} \\
\rowcolor{gray}
S386 & 0.286 & 0.638 & 0.423 & 0.649 & 0.061 & 0.780 & \textbf{0.021} & \textbf{0.896} & 0.024 & 0.862 \\
S392 & 0.340 & 0.590 & 0.414 & 0.580 & 0.067 & 0.804 & 0.038 & \textbf{0.868} & \textbf{0.037} & 0.861 \\
\rowcolor{gray}
S393 & 0.121 & 0.689 & 0.177 & 0.649 & 0.117 & 0.751 & 0.088 & \textbf{0.865} & \textbf{0.086} & 0.818 \\
S394 & 0.199 & 0.686 & 0.239 & 0.685 & 0.141 & 0.774 & 0.117 & \textbf{0.864} & \textbf{0.108} & 0.837 \\
\midrule
Avg & 0.242 & 0.649 & 0.318 & 0.636 & 0.079 & 0.795 & 0.053 & \textbf{0.878} & \textbf{0.048} & 0.863 \\
\bottomrule
\end{tabular}
\label{tab:mocap-geomtry}
}
\end{table}

\begin{table}[!t]
\caption{
\textbf{Frame Consistency on ZJU-Mocap~\cite{peng2021neural}}. 
\cj{
We report the optical flow scores to measure the video consistencies for different models. 
[$x$, $y$] indicates a full model trained with $L_{ind}=x$ and $L_{d}=y$ while {[}a{]} represents a model trained only with the pose-dependent branch and $L_{d}=a$.
Our method \textcolor{red}{{[}5, 10{]}} achieves notably better consistency.
}
}
\centering
\vspace{0.5em}
\resizebox{0.95\linewidth}{!}{
\setlength{\tabcolsep}{3pt}

\begin{tabular}{ccccccccccc}
\toprule
HumanNeRF & MonoHuman & NPC   & Vid2Avatar & {[}5{]} & {[}6{]} & {[}8{]} & {[}10{]} & \textcolor{red}{{[}5, 10{]}}  & {[}6, 10{]} & {[}8, 10{]} \\
\midrule
0.170 & 0.162 & 0.159 & 0.163 & 0.167 & 0.164 & 0.159 & 0.163 & \textbf{0.154} & 0.160 & 0.166 \\
\bottomrule
\end{tabular}

\vspace{-2.5em}
\label{tab:mocap-consistency}
}
\end{table}

\begin{table*}[b]
\caption{
\cj{
\textbf{ Ablation studies on frequency assignments.} 
Together with Tab.~\ref{tab:mocap-consistency}, our full model outperforms all ablated baselines across all metrics, proving the importance of our frequency assumptions.
}
}
\centering
\resizebox{0.9\linewidth}{!}{
\setlength{\tabcolsep}{3pt}
\begin{tabular}[b]{ccccccc}
\toprule
& \multicolumn{3}{c}{Novel view} & \multicolumn{3}{c}{Novel pose} \\
\cmidrule(lr){2-4}\cmidrule(lr){5-7}
 & PSNR$\uparrow$ & SSIM$\uparrow$ & LPIPS$\downarrow$  & PSNR$\uparrow$ & SSIM$\uparrow$ & LPIPS$\downarrow$ \\
\midrule
{[}5{]} & 30.24 & 0.966 & 30.78 & 29.58 & 0.964 & 30.34 \\
\rowcolor{gray}
{[}6{]} & 30.14 & 0.965 & 30.74 & 29.46 & 0.963 & 30.40 \\
{[}8{]} & 30.27 & 0.965 & 30.89 & 29.53 & 0.963 & 30.02 \\
\rowcolor{gray}
{[}10{]} & 30.13 & 0.965 & 31.00 & 29.36 & 0.962 & 31.01 \\
{[}6, 6{]} & 29.92 & 0.963 & 37.32 & 29.56 & 0.963 & 35.46 \\
\rowcolor{gray}
{[}6, 8{]} & 29.88 & 0.964 & 40.14 & 29.52 & 0.964 & 38.79 \\
{[}8, 8{]} & 30.11 & 0.963 & 37.98 & 29.53 & 0.963 & 35.53 \\
\rowcolor{gray}
{[}6, 10{]} & 29.91 & 0.964 & 35.25 & 29.55 & 0.964 & 33.58 \\
{[}8, 10{]} & 29.18 & 0.958 & 56.72 & 28.76 & 0.957 & 54.61 \\
\rowcolor{gray}
{[}10, 10{]} & 29.86 & 0.963 & 35.30 & 29.44 & 0.962 & 34.23 \\
\midrule
\textbf{{[}5, 10{]}$_{(\mathrm{Ours})}$}& \textbf{30.36}& \textbf{0.967} & \textbf{30.03}& \textbf{29.68}& \textbf{0.965}& \textbf{29.48}\\
\midrule
\end{tabular}%
\label{tab:supp_ablation_freq}%
}%
\end{table*}%

\begin{table*}[t]
\caption{\textbf{Evaluating importance of training Vid2Avatar with mask.} 
Extracting masks for training, Vid2Avatar produces better quantitative metrics across almost all subjects in all three metrics.
V2A$^*$ and V2A$_{mask}$ denote the original Vid2Avatar model and mask enhanced model respectively.
}
\centering
\resizebox{\linewidth}{!}{
\setlength{\tabcolsep}{0pt}
\begin{tabular}{lcccccccccccccccccccccccc}
\toprule
& \multicolumn{3}{c}{S313} & \multicolumn{3}{c}{S377} & \multicolumn{3}{c}{S386} & \multicolumn{3}{c}{S387} & \multicolumn{3}{c}{S390} & \multicolumn{3}{c}{S392} & \multicolumn{3}{c}{S393} & \multicolumn{3}{c}{S394} \\
\cmidrule(lr){2-4}\cmidrule(lr){5-7}\cmidrule(lr){8-10}\cmidrule(lr){11-13}\cmidrule(lr){14-16}\cmidrule(lr){17-19}\cmidrule(lr){20-22}\cmidrule(lr){23-25}%
  & PSNR$\uparrow$  & SSIM$\uparrow$  & LPIPS$\downarrow$  & ~PSNR$\uparrow$& SSIM$\uparrow$  & LPIPS$\downarrow$  & ~PSNR$\uparrow$& SSIM$\uparrow$  & LPIPS$\downarrow$  & ~PSNR$\uparrow$& SSIM$\uparrow$  & LPIPS$\downarrow$  & ~PSNR$\uparrow$& SSIM$\uparrow$  & LPIPS$\downarrow$  & ~PSNR$\uparrow$& SSIM$\uparrow$  & LPIPS$\downarrow$  & ~PSNR$\uparrow$& SSIM$\uparrow$  & LPIPS$\downarrow$  & ~PSNR$\uparrow$& SSIM$\uparrow$  & LPIPS$\downarrow$ \\
\midrule
\multicolumn{25}{l}{\textbf{Novel View Synthesis}}\\
V2A$^*$ & 26.82 & 0.913 & 98.74 & 29.51 & 0.951 & 70.62 & 31.16 & 0.944 & 76.46 & 28.09 & 0.938 & 83.55 & 28.07 & 0.923 & 92.48 & 30.37 & 0.943 & 82.93 & 27.19 & 0.929 & 83.42 & 28.95 & 0.931 & 81.31 \\
\rowcolor{gray}
V2A$_{mask}$ & \textbf{28.44} & \textbf{0.966} & \textbf{37.74} & \textbf{29.52} & \textbf{0.975} & \textbf{24.91} & \textbf{33.04} & \textbf{0.977} & \textbf{30.65} & \textbf{28.15} & \textbf{0.965} & \textbf{40.26} & \textbf{30.73} & \textbf{0.970} & \textbf{37.33} & \textbf{30.46} & \textbf{0.971} & \textbf{36.58} & \textbf{27.94} & \textbf{0.962} & \textbf{40.93} & \textbf{29.82} & \textbf{0.965} & \textbf{36.46} \\
\midrule
\multicolumn{25}{l}{\textbf{Novel Pose Rendering}}\\
V2A$^*$ & 26.67 & 0.913 & 97.79 & \textbf{29.86} & 0.953 & 68.45 & 30.85 & 0.941 & 77.00 & \textbf{27.42} & 0.938 & 80.78 & 28.88 & 0.926 & 88.08 & 30.20 & 0.942 & 84.57 & 27.20 & 0.929 & 82.50 & 28.54 & 0.932 & 78.80 \\
\rowcolor{gray}
V2A$_{mask}$ & \textbf{28.00} & \textbf{0.965} & \textbf{38.43} & 29.80 & \textbf{0.977} & \textbf{24.77} & \textbf{32.48} & \textbf{0.975} & \textbf{32.43} & 27.34 & \textbf{0.964} & \textbf{40.83} & \textbf{30.81} & \textbf{0.971} & \textbf{34.73} & \textbf{30.50} & \textbf{0.971} & \textbf{37.53} & \textbf{28.09} & \textbf{0.962} & \textbf{40.17} & \textbf{29.24} & \textbf{0.963} & \textbf{36.61} \\
\midrule
\end{tabular}
\label{tab:v2a_mask}
}
\end{table*}

\begin{figure*}[t!]
    \centering
    \includegraphics[width=0.9\linewidth,trim={0 20cm 18cm 0cm},clip]{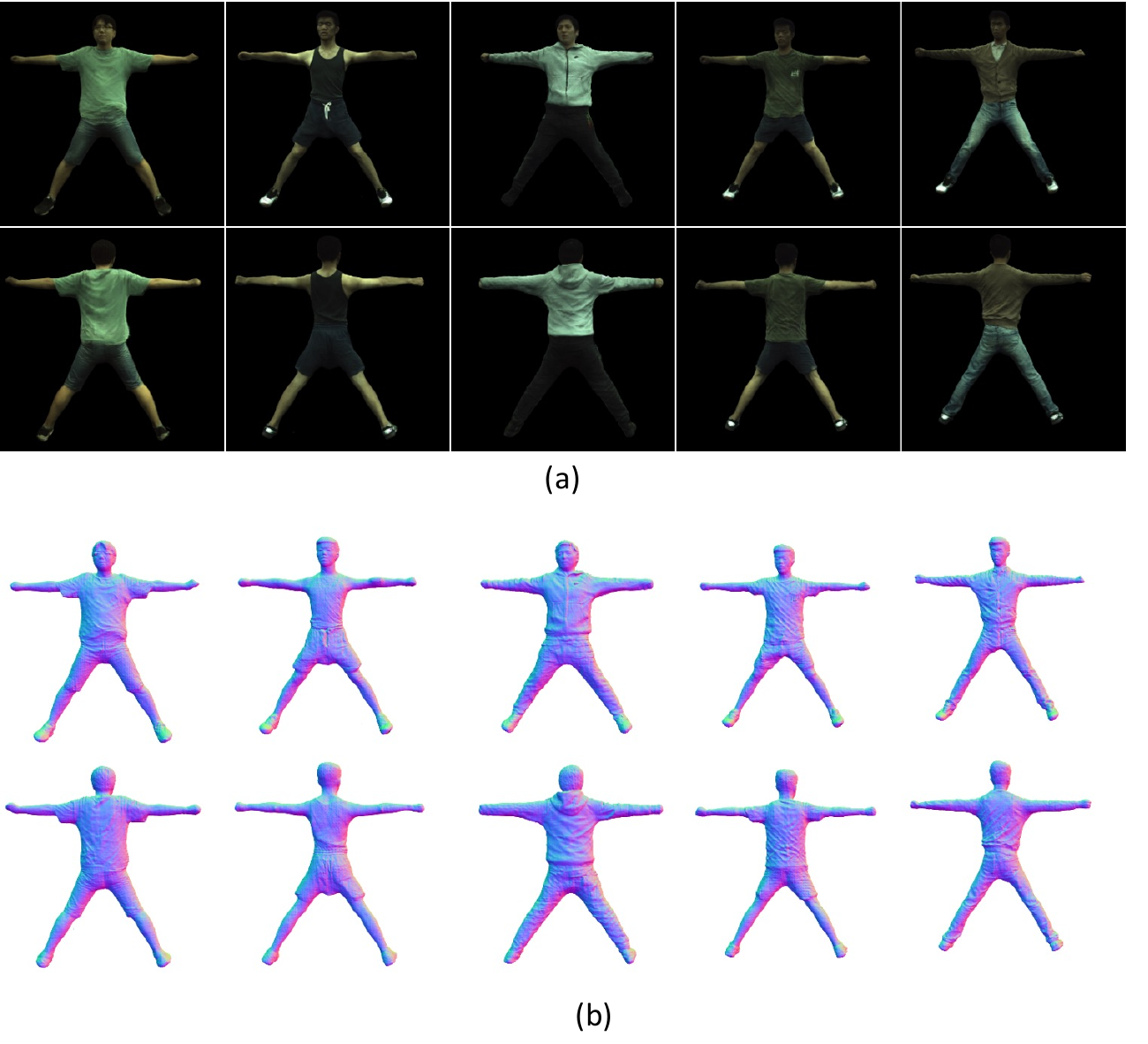}
    \caption{
    Canonical appearances (a) and normal maps (b) for ZJU-MoCap sequences.
    }
    \label{fig:canonical_info}
\end{figure*}

\section{Canonical Appearances}
\label{sec:supp_can_appear}

Learning an expressive canonical space of avatar representations is a key part for de\-for\-mation-based human modeling.
Thus we visualize the front and back
view of our reconstructed 3D avatars in canonical space in Fig.~\ref{fig:canonical_info}.
All results are rendered on the ZJU-MoCap dataset~\cite{peng2021neural}.
Our canonical images successfully capturs both the large-scale outline patterns and fine-grained texture details.

\section{Training Vid2Avatar with Masks}
\label{sec:supp_v2a}

\begin{figure*}[t!]
    \centering
    \includegraphics[width=0.95\linewidth,trim={0 30cm 0cm 0cm},clip]{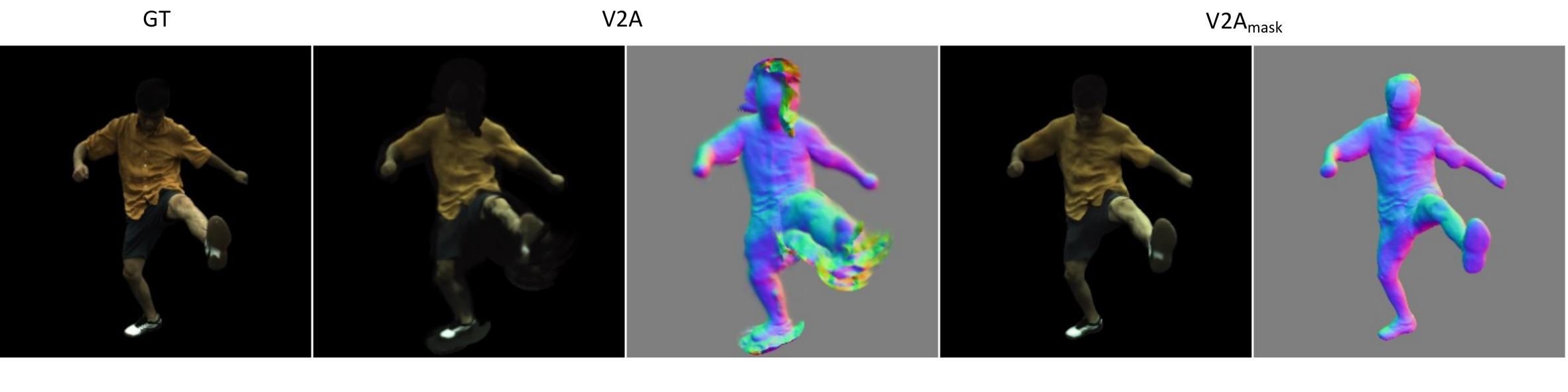}
    \caption{
    \textbf{Training Vid2Avatar with Masks.}
    The extracted masks significantly improve the original Vid2Avatar's capability in preserving stable contours and avoiding artifacts.
    }
    \label{fig:v2a_mask}
\end{figure*}

The original Vid2Avatar paper asserts that the simultaneous learning of separating humans from any background and reconstructing intricate avatar surfaces is pivotal. However, in ZJU-Mocap sequences, where the background is predominantly black and certain foreground elements appear dark, the task of precisely extracting the foreground and synthesizing texture details becomes notably more challenging.
To better compare with this SDF-based volume rendering baseline, we thus impose the ground truth supervision and explicitly extract the foreground by applying the mask for network training. 
We denote the original version and the mask enhanced version as \textbf{V2A$^*$} and \textbf{V2A$_{mask}$} respectively and list their performance on ZJU-MoCap dataset in Tab.~\ref{tab:v2a_mask}.
We can see that the mask enhanced model constantly outperforms the original implementation across all sequences, proving our configuration effectiveness.
Fig.~\ref{fig:v2a_mask} shows extra visual results to comply with the aforementioned findings.
Due to the overall superior qualitative and quantitative outputs from the mask enhanced Vid2Avatar, we only report the results of \textbf{V2A$_{mask}$} in the main text.

\section{Limitation and Discussions}
\label{sec:supp_discussion}

\begin{figure*}[t!]
    \centering
    \includegraphics[width=0.95\linewidth,trim={0 28cm 0cm 0cm},clip]{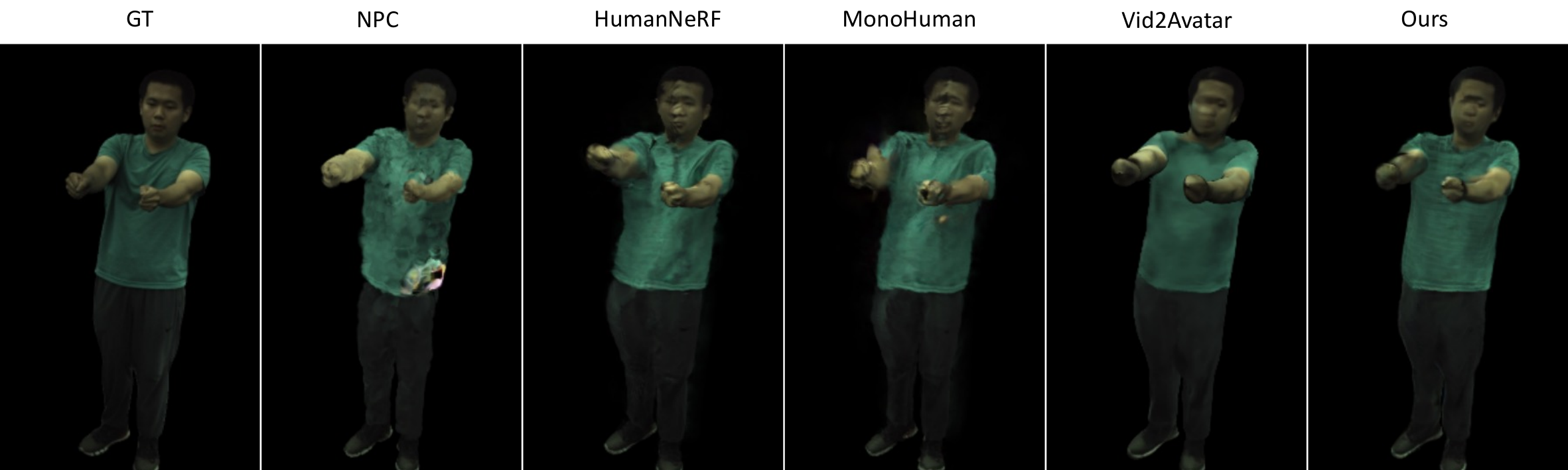}
    \caption{
    \textbf{Failure cases.}
    How to generalize to challenging cases is still an open problem, where all methods fail to preserve the visually pleasing textures and body outlines under this pose.
    }
    \label{fig:failure_case}
\end{figure*}

Due to the dense MLP computation in the volume rendering framework, computation time remains a constraint for real-time applications. 
To address this issue, some works apply the grid-based implementation to constrain the representation computation in a local area~\cite{muller2022instant,chen2022tensorf,wu2022neural}.
Our method requires individual training for each actor and cannot generalize to other humans without additional training. 
Training a generalizable human representation with foundation models is a promising direction.
Since we do not explicitly consider pattern editing in our current framework, how to enhance it with editing features is also our future work.
In Fig.~\ref{fig:failure_case}, we demonstrate that, under extreme challenges, when the test pose is very different from the training poses, our method cannot fully adaptively reproduce the target textures but distorts the body contours. 
However, it is important to recognize that addressing these issues remains an open problem in neural avatar modeling.

\paragraph{Social Impacts.}
Our research offers the potential for significantly enhancing the efficiency of human avatar modeling pipelines, thereby fostering inclusivity for underrepresented individuals and activities within supervised datasets. Nevertheless, it's crucial to confront the ethical dimensions and potential hazards associated with generating 3D models without consent. Users should utilize datasets expressly collected for motion capture algorithm development, adhering to appropriate consent and ethical guidelines.


\end{document}